\documentclass[10pt,twocolumn,letterpaper]{article}

\usepackage{cvpr}              %

\definecolor{cvprblue}{rgb}{0.21,0.49,0.74}
\usepackage[pagebackref,breaklinks,colorlinks,allcolors=cvprblue]{hyperref}

\usepackage{algorithm}
\usepackage{algpseudocode}

\usepackage{multirow}
\usepackage{booktabs}
\usepackage{array}
\usepackage{graphicx}   %
\usepackage{makecell}   %
\usepackage[table]{xcolor}
\usepackage{placeins}
\usepackage{afterpage}

\usepackage[T1]{fontenc}
\usepackage[utf8]{inputenc}
\usepackage{textcomp}

\usepackage{adjustbox}

\def\paperID{} %
\def\confName{CVPR}
\def\confYear{2026}

\title{
\textcolor[HTML]{C52C47}{G}%
\textcolor[HTML]{D53E4F}{e}%
\textcolor[HTML]{F69B5E}{o}%
\textcolor[HTML]{DCCE70}{W}%
\textcolor[HTML]{A9D67A}{o}%
\textcolor[HTML]{75C08F}{r}%
\textcolor[HTML]{4EA6A9}{l}%
\textcolor[HTML]{3288BD}{d}:
Geometric World Models}

\author{Zeyu Zhang$^{1}$\quad
Danning Li$^{2}$\quad
Ian Reid$^{2}$\quad
Richard Hartley$^{1}$\\[0.2cm]
$^1$ANU\quad
$^2$MBZUAI\\[0.2cm]
{\tt\small\bfseries \url{https://steve-zeyu-zhang.github.io/GeoWorld}}
}

\begin{document}
\makeatletter
\let\@oldmaketitle\@maketitle%
\renewcommand{\@maketitle}{\@oldmaketitle%
\includegraphics[width=\linewidth]{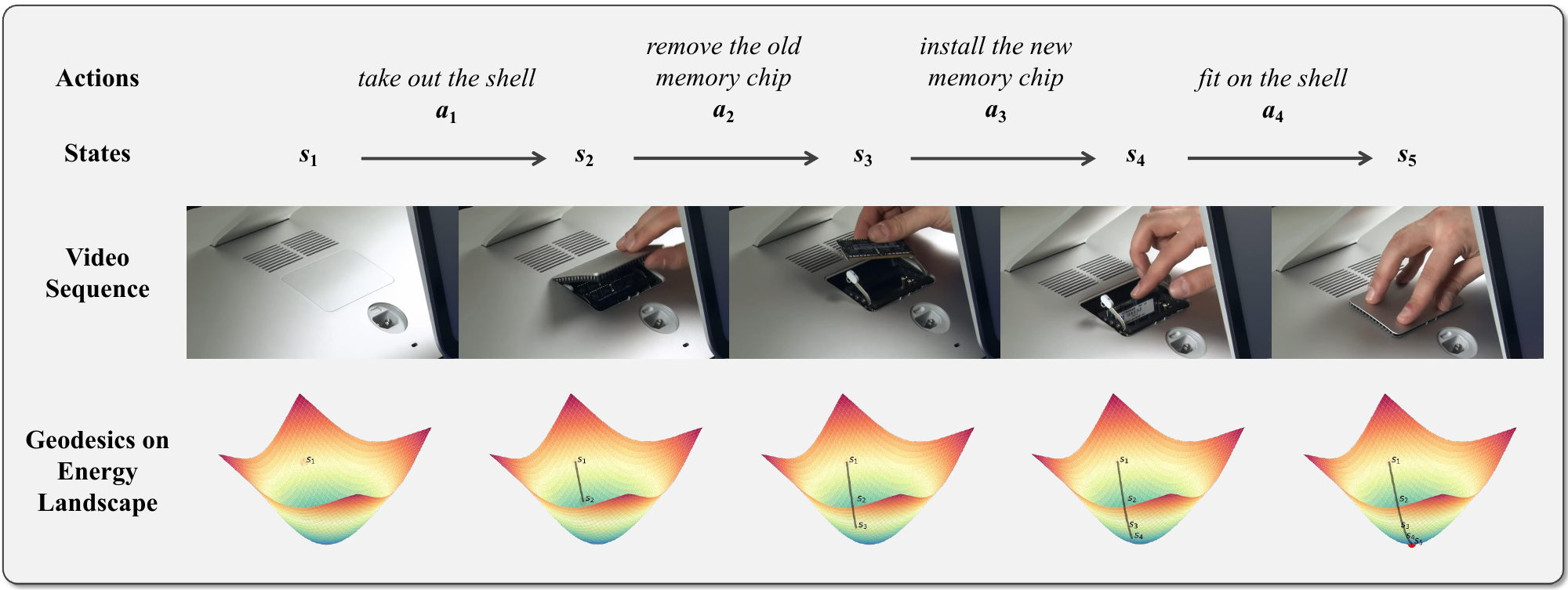}
\captionof{figure}{\textbf{Energy-based planning by GeoWorld.} 
The diagram shows a \textit{Replace Memory Chip} task from the COIN dataset~\cite{tang2019coin}, where GeoWorld plans actions by following geodesics over a hyperbolic energy landscape rather than generating pixels.}
\label{fig:teaser}
\bigskip}%
\makeatother
\maketitle
\begin{abstract}
    Energy-based predictive world models provide a powerful approach for multi-step visual planning by reasoning over latent energy landscapes rather than generating pixels. However, existing approaches face two major challenges: (\textit{i}) their latent representations are typically learned in Euclidean space, neglecting the underlying geometric and hierarchical structure among states, and (\textit{ii}) they struggle with long-horizon prediction, which leads to rapid degradation across extended rollouts.
    To address these challenges, we introduce \textbf{GeoWorld}, a geometric world model that preserves geometric structure and hierarchical relations through a \textit{Hyperbolic JEPA}, which maps latent representations from Euclidean space onto hyperbolic manifolds. We further introduce \textit{Geometric Reinforcement Learning} for energy-based optimization, enabling stable multi-step planning in hyperbolic latent space. Extensive experiments on CrossTask and COIN demonstrate around \textbf{3\%} SR improvement in 3-step planning and \textbf{2\%} SR improvement in 4-step planning compared to the state-of-the-art V-JEPA~2.
\end{abstract}

\section{Introduction}

Autoregressive (AR) next-token prediction has endowed large language models (LLMs)~\cite{yang2025qwen3} and vision-language models (VLMs)~\cite{openai2025gpt5systemcard,comanici2025gemini} with extensive world knowledge and reasoning capability, enabling them to effectively tackle complex tasks involving searching~\cite{wei2022chain}, reasoning~\cite{guo2025deepseek,huang20253d,liu2025nav}, and planning~\cite{black2024pi0,intelligence2025pi05,ye2025vla,song2025maniplvm}. Although the success of LLMs stems from modeling within language space, which serves as a shortcut toward human level knowledge~\cite{mitchell2023debate}, they still fail to fully represent the rich information of the real world, such as its physical and geometric properties~\cite{assran2025v}. In the real world, human and biological cognition often acquire knowledge primarily through visual information rather than relying solely on language, as vision offers a higher bandwidth of information than language~\cite{sigman2008brain}. For example, human infants learn mainly from visual perception during the first few months before developing a language system~\cite{kellman2006infant}, and some animals do not possess language at all~\cite{cheney1998animals}. Therefore, there are world models~\cite{mendonca2023structured,ren2025videoworld,singh2025hand,bardes2024revisiting,assran2025v} that learn solely from visual input, such as videos, and perform planning with either a generative or a predictive approach. Generative world models~\cite{mendonca2023structured,ren2025videoworld,singh2025hand} explicitly generate pixels or latent visual tokens that decode into pixels in order to predict only one step at a time~\cite{spong2002adaptive}. As a result, they lack awareness of the full trajectory structure or the energy landscape over multiple steps. In contrast, predictive world models~\cite{assran2023self,garrido2024learning,bardes2024revisiting,assran2025v} such as JPEA~\cite{lecun2022path} do not generate pixels. Instead, they learn an energy landscape in latent space that measures the compatibility between current and target states. This enables multi-step hierarchical planning, where high-level reasoning minimizes energy in latent space, while lower-level modules fill in the physical details.

However, existing energy-based predictive world models face two significant challenges: 

\textbf{(1) Geometric neglect.} Although predictive world models perform multi-step hierarchical planning in latent space, their representations are typically learned in a Euclidean space without preserving the underlying geometric relations among states.
As a result, the learned energy landscape fails to capture meaningful geodesic distances or hierarchical embeddings between latent states~\cite{nickel2017poincare}, which weakens the model’s ability to perform geometry-consistent planning over long horizons.

\textbf{(2) Multi-step shortcoming.}
Multi-step videos are limited and expensive to acquire, so existing predictive world models are primarily trained on one-step video transitions~\cite{goyal2017something,kuehne2011hmdb,soomro2012ucf101,caba2015activitynet,kay2017kinetics,miech2019howto100m}. Although learning an energy landscape over entire trajectories conceptually enables long-horizon planning, their performance degrades rapidly as the planning horizon increases, exposing a weakness in modeling long-term temporal dependencies.

Our motivation is to address these problems from a \textit{geometric} perspective. For the first challenge, a geometry-aware world model is required to preserve geometric properties when learning the energy landscape for hierarchical planning. 
For the second challenge, reinforcement learning (RL) has proven effective in adjusting a pretrained foundation model when its outputs are unsatisfactory in certain aspects~\cite{ouyang2022training,schulman2017proximal}. 
Therefore, a geometry-aware RL method is required to obtain optimal trajectories on the latent manifold, improving the model’s multi-step planning capability.

\begin{figure}[t]
    \centering
    \begin{subfigure}{0.48\linewidth}
        \centering
        \includegraphics[width=\linewidth]{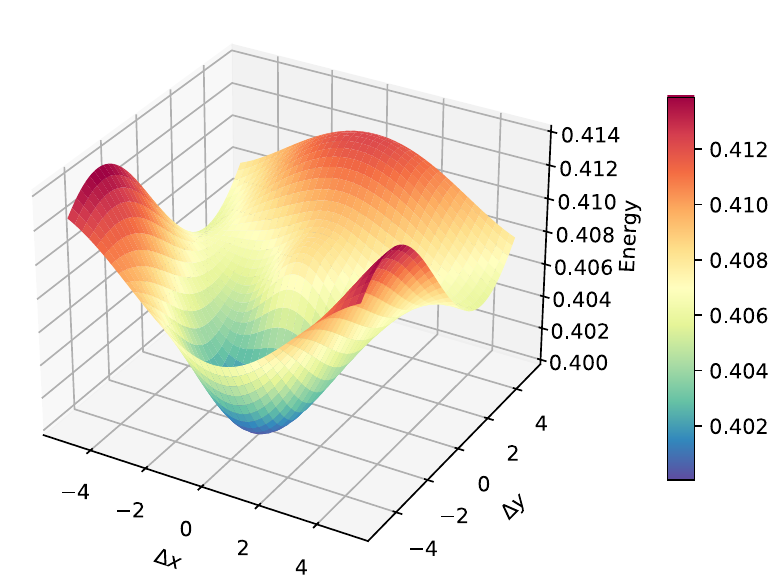}
        \caption{V-JEPA 2~\cite{assran2025v} Energy Landscape}
        \label{fig:energy_vjepa2}
    \end{subfigure}
    \hfill
    \begin{subfigure}{0.48\linewidth}
        \centering
        \includegraphics[width=\linewidth]{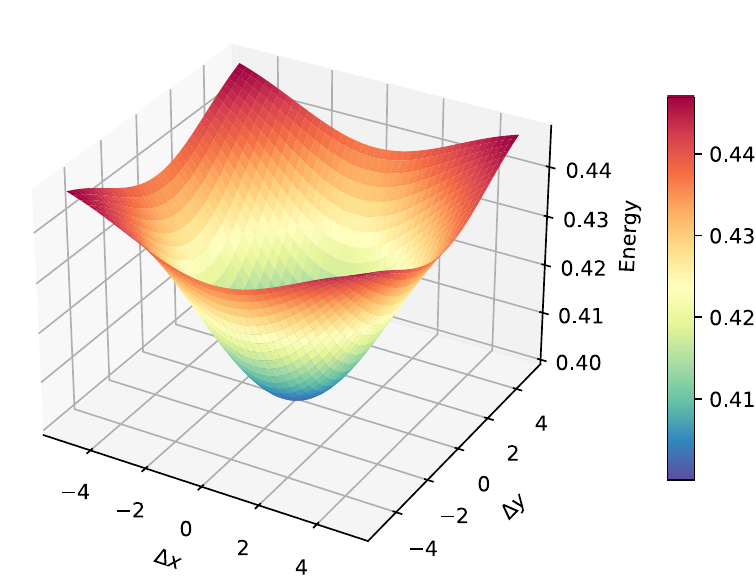}
        \caption{\textbf{GeoWorld} Energy Landscape}
        \label{fig:energy_geoworld}
    \end{subfigure}

    \caption{
    \textbf{Energy landscape comparison for V-JEPA 2~\cite{assran2025v} and GeoWorld.}
    We visualize the energy by sweeping two orthonormal tangent-space directions
    $(\Delta x, \Delta y)$ around a reference latent state.
    GeoWorlds yields a structured, curvature-aware energy landscape that better reflects geometric structure and hierarchical relations among latent states and improves energy-based planning. For more details see \textit{Appendix}~\ref{sec:energy_landscape}.}
    \label{fig:energy_landscape}
\end{figure}

Hence, we introduce the \textbf{Geometric World Model} (\textbf{GeoWorld}), a method that enhances energy-based predictive world models by preserving geometric structure and hierarchical awareness in latent space, as shown in Figure~\ref{fig:teaser}.
To address the first challenge, we propose \textit{Hyperbolic JEPA} (\textit{H-JEPA}), which maps latent representations from Euclidean space $\mathbb{R}^n$ onto a hyperbolic manifold $\mathbb{H}^n$, where geodesic distances naturally encode hierarchical relations among states. 
By learning dynamics along hyperbolic geodesics, H-JEPA preserves latent geometry during multi-step prediction, ensuring that the learned energy landscape aligns with the underlying structure of the physical world and supports geometry-consistent planning, as shown in \ref{fig:energy_landscape}.

To address the second challenge, we design a \textit{Geometric Reinforcement Learning} (\textit{GRL}) that reformulates multi-step planning as the optimization of an energy-based value function, where lower hyperbolic energy corresponds to higher cumulative reward. GRL directly optimizes the predictor of the world model without training an additional policy or reward model. 
By adjusting the predictor’s energy-based value representation through hyperbolic geodesics minimization and triangle inequality regularization, GRL enforces geodesic-consistent rollouts on the latent manifold, effectively improving long-horizon stability and planning performance.

To verify our method’s capability on long-horizon planning, we evaluate multi-step goal-conditioned visual planning on standard benchmarks, including CrossTask~\cite{zhukov2019cross} and COIN~\cite{tang2019coin}. Our GeoWorld achieves consistent improvements over the previous state-of-the-art predictive world model V-JEPA~2, including improvements of around \textbf{3\% SR} in 3-step planning and \textbf{2\% SR} in 4-step planning across both datasets.

The contributions of our work can be summarized as follows:
\begin{itemize}
    \item We introduce the \textbf{Geometric World Model} (\textbf{GeoWorld}) with a \textit{Hyperbolic JEPA} (\textit{H-JEPA}), which preserves geometric structure and hierarchical relations by mapping latent representations onto a hyperbolic manifold and learning dynamics along hyperbolic geodesics, resulting in a geometry-consistent energy landscape for multi-step prediction and planning.
    
    \item We propose \textit{Geometric Reinforcement Learning} (\textit{GRL}), an energy-based optimization framework that directly refines the predictor through hyperbolic energy minimization and triangle-inequality regularization, enabling geodesic-consistent rollouts and improving long-horizon planning stability.

    \item We demonstrate strong performance on long-horizon goal-conditioned visual planning across CrossTask and COIN, achieving around \textbf{3\%} SR improvement in 3-step planning and \textbf{2\%} SR improvement in 4-step planning compared to V-JEPA~2.
\end{itemize}

\section{Related Works}

\paragraph{Video World Models}
There are two primary approaches for video world modeling: generative world models~\cite{mendonca2023structured,ren2025videoworld,singh2025hand} and predictive world models~\cite{lecun2022path,assran2023self,garrido2024learning,bardes2024revisiting,assran2025v}. Generative world models typically build upon autoregressive~\cite{deng2024autoregressive,wang2024loong,liu2025infinitystar} or semi-autoregressive~\cite{ren2025autoregressive,huang2025self,cui2025self,teng2025magi,yang2025longlive,chen2025skyreels} architectures that observe the visual context and explicitly generate the next frame or its latent representation. These models often incorporate an inverse dynamics module~\cite{spong2002adaptive} trained to infer actions from consecutive observations, enabling one-step reactive control but preventing multi-step reasoning because the model lacks access to the global trajectory structure and cannot capture long-range dynamics. Moreover, generative approaches must decode visual tokens or pixels during planning, which introduces unnecessary noise and computational overhead and limits their ability to model abstract energy landscapes for hierarchical planning~\cite{ren2025videoworld}. In contrast, predictive world models do not generate pixels. Instead, they learn an energy landscape in latent space that quantifies the compatibility between current and target states~\cite{lecun2022path}. This design allows for multi-step trajectory optimization using sampling-based planners such as the cross-entropy method (CEM)~\cite{de2005tutorial}, enabling long-horizon planning without explicit pixel decoding.

\paragraph{Goal-Conditioned Visual Planning}
\label{sec:goal}

Goal conditioned visual planning aims to produce a sequence of actions that achieves a given goal based on visual observations. Prior works have evolved into three independent setups depending on the modalities of the observation and the goal, which may be images, videos, or language. (1) In visual planning for assistance (VPA)~\cite{patel2023pretrained}, observations are videos and goals are described in natural language. This typically requires models built on LLMs with multimodal processing capability~\cite{islam2024propose,zhang2025enhancing,chen2025planning}. (2) In procedural planning (PP)~\cite{chang2020procedure}, both observations and goals are specified as images without any language involved, which limits the model’s ability to capture temporal information in the physical world~\cite{liu2023language,islam2024propose,niu2024schema,bi2021procedure,zhao2022p3iv,wang2023pdpp,nagasinghe2024not,shi2025actiondiffusion,zhou2025masked,sun2022plate,wang2023event}. (3) In visual planning with videos, both observations and goals are given as videos, which aligns more naturally with temporal dynamics in the real world and is commonly addressed using video LLMs~\cite{wang2025internvl3_5,yang2025qwen3,comanici2025gemini,openai2025gpt5systemcard}, generative world models~\cite{ren2025videoworld}, and predictive world models~\cite{assran2025v}.

\begin{figure*}[t]
    \centering
    \includegraphics[width=\linewidth]{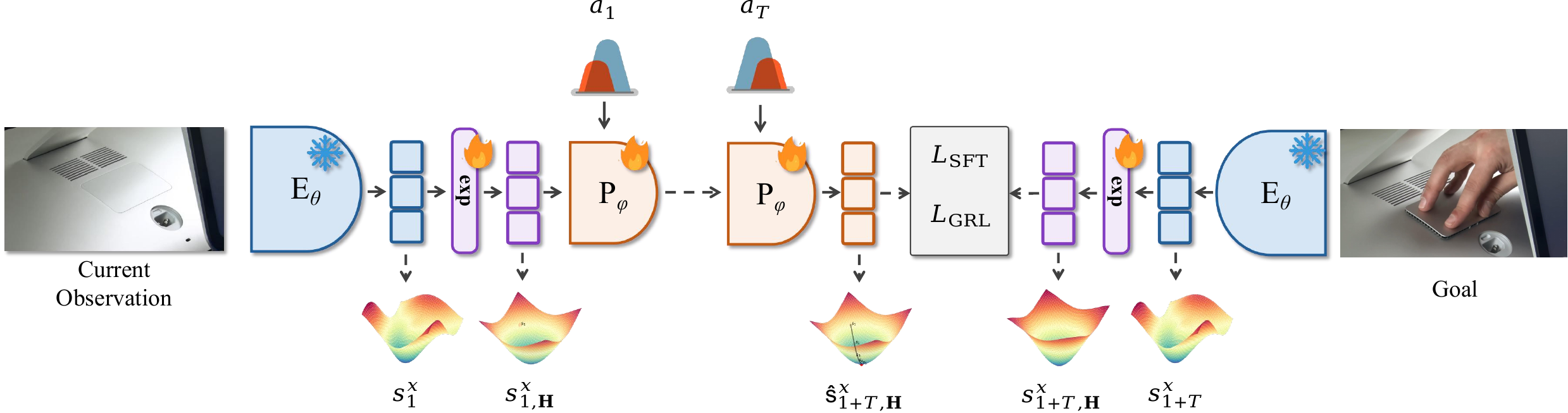}
    \caption{\textbf{Overview of GeoWorld.} Our geometric world model integrates Hyperbolic JEPA for geometry-preserving latent dynamics and Geometric Reinforcement Learning for geodesic-consistent multi-step refinement. Together with energy-based planning using CEM, GeoWorld enables stable and geometry-aware long-horizon visual planning.}
    \label{fig:architecture}
\end{figure*}

\section{Method}

\subsection{Overview}

We introduce \textbf{GeoWorld}, a geometric world model designed to enhance long-horizon visual planning by preserving geometric structure and hierarchical awareness in latent space. To address the limitation of Euclidean latent representations, GeoWorld incorporates \textit{Hyperbolic JEPA} (H-JEPA), which maps encoder outputs from Euclidean space onto a hyperbolic manifold where geodesic distances naturally encode hierarchical relations among states. By learning latent dynamics along hyperbolic geodesics, H-JEPA enforces geometry-consistent transitions that better reflect the structure of real-world trajectories. To further improve stability in multi-step prediction, we develop \textit{Geometric Reinforcement Learning} (GRL), an energy-based optimization framework that treats planning as minimizing a hyperbolic value function without training an additional policy or reward model. GRL refines the predictor through hyperbolic energy minimization and triangle-inequality regularization, encouraging geodesic-consistent rollouts and improving long-horizon temporal coherence. Leveraging energy-based planning with the Cross-Entropy Method (CEM)~\cite{de2005tutorial} further enables efficient trajectory optimization by searching for action sequences that follow geodesic paths in hyperbolic latent space. Together, H-JEPA and GRL form the core of GeoWorld, enabling geometry-aware multi-step planning in predictive world models, as shown in Figure~\ref{fig:architecture}. 

For preliminaries on JEPA~\cite{lecun2022path}, hyperbolic geometry, and the value function in RL, see \textit{Appendix}~\ref{sec:prelim}.

\subsection{Hyperbolic JEPA}

From a representation perspective, we aim to learn a mapping from states onto a hyperbolic space $\mathbb{H}^n$ such that the optimal plan corresponds to a geodesic in hyperbolic space.
Hence, we propose \textit{Hyperbolic JEPA} (\textit{H-JEPA}), which models latent dynamics on the hyperbolic manifold to preserve hierarchical relations and underlying geometric coherence during multi-step planning.

We define the observation at time $t$ as $x_t$. 
$E_\theta(\cdot)$ denotes the pretrained encoder~\cite{assran2025v}, which encodes the observation $x_t$ into the latent state $s_t^{x}$:
\begin{equation}
s_t^{x} = E_\theta(x_t) \in \mathbb{R}^n.
\end{equation}

To effectively map the encoder output from Euclidean space $\mathbb{R}^n$ to hyperbolic space $\mathbb{H}^n$, 
we interpret the Euclidean embedding $s_t^{x}$ as a tangent vector in the tangent space $\mathbf{T}_0\mathbb{H}^n$ at the origin. 
We then apply the exponential map at the origin of the Poincaré ball model $\mathbb{B}_c^n$ with curvature $K = -c$, 
which projects the tangent vector onto the hyperbolic manifold, as detailed in \textit{Appendix}~\ref{sec:poincare}. 

Formally, the hyperbolic latent state is obtained as
\begin{equation}
s_{t,\mathbb{H}}^{x} = \exp_0(s_t^{x}) 
= \tanh\!\big(\sqrt{c}\|s_t^{x}\|\big)\frac{s_t^{x}}{\sqrt{c}\|s_t^{x}\|},
\quad s_{t,\mathbb{H}}^{x} \in \mathbb{B}_c^n.
\end{equation}

Then the action-conditioned predictor $P_\phi(\cdot)$ takes a sequence of hyperbolic latent states $(s_{t,\mathbb{H}}^{x})_{t=1}^{T}$ 
and a corresponding sequence of actions $(a_t)_{t=1}^{T}$ as input, and predicts the sequence of next-state representations 
$(\hat{s}_{t+1,\mathbb{H}}^{x})_{t=1}^{T}$ over a planning horizon $T$:
\begin{equation}
(\hat{s}_{t+1,\mathbb{H}}^{x})_{t=1}^{T}
= 
P_{\phi}\!\big((s_{t,\mathbb{H}}^{x}, a_t)_{t=1}^{T}\big).
\end{equation}

and $\theta$ and $\phi$ denote the parameters (weights) of the encoder and predictor networks, respectively.

\subsection{Training Objective}

The supervised training objective of H-JEPA is to learn a predictive world model that follows the geodesic path of minimum energy cost between the current and target latent states in hyperbolic space. 
Specifically, the model minimizes the Poincaré-ball hyperbolic distance $d_{\mathbb{H}}$ between the predicted and true latent representations, as defined in Eq.~\ref{eq:poincare_distance_c} of \textit{Appendix}~\ref{sec:poincare}, ensuring that each transition aligns with the lowest-energy trajectory on the manifold.

The objective consists of a joint loss combining a teacher-forcing loss and a rollout loss. 
The teacher-forcing loss encourages accurate one-step prediction by aligning the predicted next-state representation with the ground-truth latent embedding, while the rollout loss recursively feeds the model’s own predictions as inputs to enforce temporal consistency across multiple future steps. 

\paragraph{Teacher Forcing}

The teacher-forcing loss trains the model to accurately perform one-step future prediction by minimizing the hyperbolic geodesic distance \( d_{\mathbb{H}} \) between the predicted latent representation \(\hat{s}_{t+1,\mathbb{H}}^{x}\) and the encoded ground-truth latent \(s_{t+1,\mathbb{H}}^{x}\) at each time step \(t\):

\begin{align}
\mathcal{L}_{\text{TF}}(\theta,\phi)
&= 
\frac{1}{T}\sum_{t=1}^{T}
d_{\mathbb{H}}\!\left(
P_{\phi}\!\big(\exp_0\!\big(E_\theta(x_t)\big), a_t \right), \nonumber\\
&\qquad\left.\exp_0\!\big(E_\theta(x_{t+1})\big)
\right) \\
&=
\frac{1}{T}\sum_{t=1}^{T}
d_{\mathbb{H}}\!\left(
\hat{s}_{t+1,\mathbb{H}}^{x},
s_{t+1,\mathbb{H}}^{x}
\right) \\
&=
\frac{1}{T}\sum_{t=1}^{T}
\frac{1}{\sqrt{c}}\,
\operatorname{arcosh}\!\left(
1 \right. \nonumber\\
&\qquad\left. {}+ 2c
\frac{\|\hat{s}_{t+1,\mathbb{H}}^{x} - s_{t+1,\mathbb{H}}^{x}\|^2}
{(1 - c\|\hat{s}_{t+1,\mathbb{H}}^{x}\|^2)(1 - c\|s_{t+1,\mathbb{H}}^{x}\|^2)}
\right)
\end{align}

\paragraph{Rollout}

The rollout loss feeds the predictor’s output back as input, enabling the model to learn multi-step future prediction. 
In this case, we design a two-step rollout loss to enhance the model’s capability for long-horizon planning:
\begin{align}
\mathcal{L}_{\text{rollout}}(\theta,\phi)
&=
\frac{1}{T}\sum_{t=1}^{T}
d_{\mathbb{H}}\!\left(
P_{\phi}\!\big(\exp_0\!\big(E_\theta(x_t)\big), a_t, a_{t+1}\big)\right., \nonumber\\
&\qquad\left.\exp_0\!\big(E_\theta(x_{t+2})\big)
\right) \\
&=
\frac{1}{T}\sum_{t=1}^{T}
d_{\mathbb{H}}\!\left(
\hat{s}_{t+2,\mathbb{H}}^{x},
s_{t+2,\mathbb{H}}^{x}
\right) \\
&=
\frac{1}{T}\sum_{t=1}^{T}
\frac{1}{\sqrt{c}}\,
\operatorname{arcosh}\!\left(
1 \right. \nonumber\\
&\qquad\left. {}+ 2c
\frac{\|\hat{s}_{t+2,\mathbb{H}}^{x} - s_{t+2,\mathbb{H}}^{x}\|^2}
{(1 - c\|\hat{s}_{t+2,\mathbb{H}}^{x}\|^2)(1 - c\|s_{t+2,\mathbb{H}}^{x}\|^2)}
\right),
\end{align}

\paragraph{Total Loss}

Hence, the total loss in the supervised stage is defined as
\begin{equation}
\mathcal{L}_{\text{SFT}}(\theta,\phi)
=
\lambda\, \mathcal{L}_{\text{TF}}(\theta,\phi)
+
(1 - \lambda)\, \mathcal{L}_{\text{rollout}}(\theta,\phi),
\end{equation}
where \(\lambda\) is a loss weighting hyperparameter. 

Together, these two components train the predictor to learn smooth, geodesically consistent trajectories that capture both short-term accuracy and long-horizon stability within the hyperbolic latent space.

\subsection{Geometric Reinforcement Learning}

We propose a \textit{Geometric Reinforcement Learning} (\textit{GRL}) approach that improves the predictor in multi-step planning by adjusting its energy-based value representation, aligning lower energy with higher expected reward.

\paragraph{Energy Cost}
Given a frozen encoder $E$ and a trainable predictor $P_{\phi}$, we define the energy cost of moving from state $s_{t,\mathbb{H}}^{x}$ to state $s_{t+1,\mathbb{H}}^{x}$ as

\begin{align}
\label{eq:energy_cost_small}
    c_t(s_{t,\mathbb{H}}^{x}, s_{t+1,\mathbb{H}}^{x})
    &= d_{\mathbb{H}}\!\left(
        P_{\phi}\!\big(\exp_0\!\big(E(x_t)\big), a_t\big),
        \exp_0\!\big(E(x_{t+1})\big)
    \right)\\
    &= d_{\mathbb{H}}(\hat{s}_{t+1,\mathbb{H}}^{x}, s_{t+1,\mathbb{H}}^{x}).
\end{align}

which ideally indicates that we aim to minimize the energy cost of moving from state $s_{t,\mathbb{H}}^{x}$ to state $s_{t+1,\mathbb{H}}^{x}$, which is identical to minimizing the geodesic distance between the predicted state $\hat{s}_{t+1,\mathbb{H}}^{x}$ and the target state $s_{t+1,\mathbb{H}}^{x}$.

\paragraph{Reward}
We then define the reward as the negative energy cost of moving between states:
\begin{equation}
    r_t(s_{t,\mathbb{H}}^{x}, a_t, s_{t+1,\mathbb{H}}^{x}) = -c_t(s_{t,\mathbb{H}}^{x}, s_{t+1,\mathbb{H}}^{x}).
\end{equation}

\paragraph{Path Value Function}
As mentioned in \textit{Appendix \ref{sec:value}}, the value function $V$ is a mathematical object that quantifies the amount of energy required for an agent to reach optimality from a given state to a target state, where lower energy corresponds to a higher expected cumulative reward. 

Hence, the path value function between the current and goal latent states, given a planning horizon $T$, is defined as the expected cumulative reward:
\begin{equation}
    V(s_{1,\mathbb{H}}^{x}, s_{1+T,\mathbb{H}}^{x})
    = \mathbb{E}_{a_{1:T} \sim \phi}\!\left[
        \sum_{t=1}^{T} \gamma^{t-1}
        r_t(s_{t,\mathbb{H}}^{x}, a_t, s_{t+1,\mathbb{H}}^{x})
    \right],
\end{equation}

where $\gamma \in [0, 1)$ is the discount factor.

Our objective is to maximize the total reward (i.e., maximize the return) such that $P_{\phi}$ follows the geodesics. Therefore, the optimal path value function maximizes the expected cumulative reward:
\begin{align}
    V^{*}(s_{1,\mathbb{H}}^{x}, s_{1+T,\mathbb{H}}^{x})
    &= \max_{\phi}\,\mathbb{E}_{a_{1:T} \sim \phi}\!\left[
        \sum_{t=1}^{T} \gamma^{t-1} \,
        r_t(s_{t,\mathbb{H}}^{x}, a_t,
    \right. \nonumber\\
    &\qquad\qquad\qquad\qquad\left.
        s_{t+1,\mathbb{H}}^{x})
    \right]. \nonumber\\
    &= \min_{\phi}\,\mathbb{E}_{a_{1:T} \sim \phi}\!\left[
    \sum_{t=1}^{T} \gamma^{t-1} \,
    d_{\mathbb{H}}(\hat{s}_{t+1,\mathbb{H}}^{x},
    \right. \nonumber\\
    &\qquad\qquad\qquad\qquad\left.
        s_{t+1,\mathbb{H}}^{x})
    \right].
\end{align}

which is equivalent to minimizing the total hyperbolic distance between the predicted and target states.

\paragraph{Triangle Inequality Regularization}

The hyperbolic geodesic distance \( d_{\mathbb{H}} \) satisfies the triangle inequality. Therefore, for any consecutive triplet in the predictor’s rollouts:
\begin{equation}
d_{\mathbb{H}}(\hat{s}_{t,\mathbb{H}}^{x}, \hat{s}_{t+2,\mathbb{H}}^{x})
\leq
d_{\mathbb{H}}(\hat{s}_{t,\mathbb{H}}^{x}, \hat{s}_{t+1,\mathbb{H}}^{x})
+ d_{\mathbb{H}}(\hat{s}_{t+1,\mathbb{H}}^{x}, \hat{s}_{t+2,\mathbb{H}}^{x}).
\end{equation}

This indicates that minimizing the sum of consecutive step distances encourages the predicted trajectory to align with the geodesic path. Hence, we introduce a regularization term:
\begin{align}
\mathcal{L}_{\Delta}
= \frac{1}{T-2} \sum_{t=1}^{T-2}
\Big[
d_{\mathbb{H}}(\hat{s}_t, \hat{s}_{t+2})
- d_{\mathbb{H}}(\hat{s}_t, \hat{s}_{t+1}) \nonumber\\
\qquad\qquad
- d_{\mathbb{H}}(\hat{s}_{t+1}, \hat{s}_{t+2})
\Big]_+.
\end{align}

This term enforces multi-step rollout consistency by encouraging predicted trajectories to satisfy hyperbolic geodesic properties.

\paragraph{Total Loss}

Hence, the total loss in Geometric Reinforcement Learning can be expressed as
\begin{equation}
\mathcal{L}_{\mathrm{GRL}}(\phi)
= \mathbb{E}_{a_{1:T} \sim \phi}\!\left[
    \sum_{t=1}^{T} \gamma^{t-1}
    d_{\mathbb{H}}(\hat{s}_{t+1,\mathbb{H}}^{x}, s_{t+1,\mathbb{H}}^{x})
\right]
+ \beta \mathcal{L}_{\Delta}.
\end{equation}

where $\beta$ is the regularization factor.

\subsection{Energy-Based Planning}

We then perform energy-based planning after training, with the frozen encoder $E$ and predictor $P$. 
The predictor serves as a world model, capable of predicting how latent representations evolve when an action sequence is applied. 
During planning, we search for an optimal action sequence that follows the geodesic path between the current and goal latent states, effectively minimizing a goal-conditioned energy cost defined in the hyperbolic latent space. 

Given the current observation $x_1$, the future target $x_{1+T}$, and the planning horizon $T$, 
we encode the current and goal observations as
\begin{equation}
s_{1,\mathbb{H}}^{x} = \exp_0(E(x_1)), \qquad 
s_{1+T,\mathbb{H}}^{x} = \exp_0(E(x_{1+T})).
\end{equation}

We then define the energy cost function $C$ based on the Poincaré geodesic distance, which measures the hyperbolic energy between the predicted and goal latent states over the planning horizon:
\begin{equation}
C((\hat{a}_t)_{t=1}^{T}; s_{1,\mathbb{H}}^{x}, s_{1+T,\mathbb{H}}^{x})
= 
d_{\mathbb{H}}\!\left(
P((\hat{a}_t)_{t=1}^{T}; s_{1,\mathbb{H}}^{x}),\,
s_{1+T,\mathbb{H}}^{x}
\right),
\label{eq:energy_cost_hyperbolic}
\end{equation}

Hence, the optimal action sequence $(a_t^{*})_{t=1}^{T}$ is obtained by minimizing this hyperbolic energy cost:
\begin{equation}
(a_t^{*})_{t=1}^{T}
= 
\arg\min_{(\hat{a}_t)_{t=1}^{T}}
d_{\mathbb{H}}\!\left(
P((\hat{a}_t)_{t=1}^{T}; s_{1,\mathbb{H}}^{x}),\,
s_{1+T,\mathbb{H}}^{x}
\right).
\end{equation}

The optimization is performed with the Cross-Entropy Method (CEM)~\cite{de2005tutorial}, 
as detailed in Algorithm~\ref{alg:cem_planning} of \textit{Appendix}~\ref{sec:v-jepa}

\begin{equation}
(a_t^{*})_{t=1}^{T} = 
\operatorname{CEM}\!\left(
x_1,\,\, x_{1+T},\, P(\cdot),\, E(\cdot),\,
T,\, N,\, K,\, I,\, \mu_0,\, \Sigma_0
\right)
\label{eq:cem}
\end{equation}
where $x_1$ is the current observation, $x_{1+T}$ is the goal observation, 
$P_{\phi}$ is the predictor, 
$E(\cdot)$ is the encoder, $T$ is the planning horizon, 
$N$ is the number of samples, 
$K$ is the number of elites, 
$I$ is the number of iterations, 
and $(\mu_0, \Sigma_0)$ denote the initial mean and covariance of the action distribution.

\section{Experiments}

\subsection{Benchmarks and Evaluation Metrics}
\paragraph{Benchmarks}
For evaluating our world model’s capability in multi-step goal-conditioned planning, we adapt two standard goal-conditioned visual planning datasets, CrossTask~\cite{zhukov2019cross} and COIN~\cite{tang2019coin}, which contain diverse fine-grained action labels and timestamps of human daily activities.

\textit{CrossTask} consists of 4.7K videos across 83 tasks, covering 105 actions, with an average of 8 actions per video. The total duration is 375h.

\textit{COIN} consists of 11,287 videos across 180 tasks, covering 778 actions, with an average of 3.9 actions per video. The total duration is 476h.

\paragraph{Metrics}

Following previous works in goal-conditioned visual planning~\cite{bi2021procedure}, we adopt three metrics for evaluation: (1) Success Rate (SR) computes whether the predicted action sequence exactly matches the ground truth sequence. (2) Mean Accuracy (mAcc) computes the average accuracy of the predicted actions at each time step. (3) Mean Intersection over Union (mIoU) quantifies the overlap between the predicted procedure and the ground truth.

\subsection{Baseline and Evaluation Protocol}
\label{sec:baseline_protocol}

We follow previous works~\cite{chang2020procedure,bi2021procedure,ren2025videoworld,assran2025v} and evaluate goal-conditioned visual planning in two setups based on the modality of the observation and the target, as discussed in Section~\ref{sec:goal}. For \textit{procedural planning}~\cite{chang2020procedure}, both observations and goals are specified as images, which is more aligned with the traditional visual planning setup. For \textit{visual planning with videos}~\cite{ren2025videoworld}, both observations and goals are specified as video clips, which more faithfully reflect the temporal–spatial information in the real world.

For both setups, evaluation is conducted over a planning horizon \(T\), where the model outputs a sequence of \(T\) actions given the observation and the goal.

In both setups, we include three categories of baselines. \textit{LLM-based} methods leverage LLMs or VLMs for reasoning and planning~\cite{liu2023language,islam2024propose,niu2024schema,wang2025internvl3_5,yang2025qwen3,comanici2025gemini,openai2025gpt5systemcard}. \textit{Generative (world) models} explicitly generate pixels or latent visual tokens that decode into pixels for planning~\cite{chang2020procedure,bi2021procedure,zhao2022p3iv,wang2023pdpp,nagasinghe2024not,shi2025actiondiffusion,zhou2025masked,ren2025videoworld}. \textit{Predictive (world) models} predict a sequence of actions without relying on pixel generation~\cite{ehsani2018let,abu2019uncertainty,srinivas2018universal,sun2022plate,wang2023event,assran2025v}.

There are two extra baselines in the procedural planning setup. \textit{Random} randomly selects an action from all actions and serves as the empirical lower bound of performance~\cite{chang2020procedure}. The \textit{Retrieval-Based} approach retrieves the nearest neighbor by minimizing the visual feature distance within the training dataset, and the action sequence associated with the retrieved neighbor is then used as the plan~\cite{zhou2025masked}.

Besides, for both V-JEPA 2~\cite{assran2025v} and our GeoWorld, we adopt frozen encoders, while for VideoWorld~\cite{ren2025videoworld} we perform full finetuning. For general VLMs~\cite{wang2025internvl3_5,yang2025qwen3,comanici2025gemini,openai2025gpt5systemcard} in visual planning with videos, all evaluations are conducted in a zero-shot setting. For more details on the baselines, see \textit{Appendix}~\ref{sec:baseline}.

\begin{table*}[t]
\centering
\small
\caption{\textbf{Goal-conditioned visual planning with images on CrossTask~\cite{zhukov2019cross} and COIN~\cite{tang2019coin} datasets.}
We evaluate multi-step planning over a horizon \(T\) under the \textit{procedural planning} setup~\cite{chang2020procedure}, where both observations and goals are specified as images.}
\label{tab:main_pp}
\resizebox{\textwidth}{!}{
\begin{tabular}{lcccccc ccccccc}
\toprule
\multirow{3}{*}{Method}
& \multicolumn{6}{c}{CrossTask Dataset~\cite{zhukov2019cross}}
& \multicolumn{6}{c}{COIN Dataset~\cite{tang2019coin}}
\\
\cmidrule(lr){2-7}\cmidrule(lr){8-13}
& \multicolumn{3}{c}{T=3}
& \multicolumn{3}{c}{T=4}
& \multicolumn{3}{c}{T=3}
& \multicolumn{3}{c}{T=4}
\\
\cmidrule(lr){2-4}\cmidrule(lr){5-7}
\cmidrule(lr){8-10}\cmidrule(lr){11-13}
& SR & mAcc & mIoU
& SR & mAcc & mIoU
& SR & mAcc & mIoU
& SR & mAcc & mIoU
\\
\midrule
Random~\cite{chang2020procedure}
& 0.01 & 0.94 & 1.66 & 0.01 & 1.83 & 1.66
& 0.01 & 0.01 & 2.47 & 0.01 & 0.01 & 2.32 \\
Retrieval-Based~\cite{zhou2025masked}
& 8.05 & 23.30 & 32.06 & 3.95 & 22.22 & 36.97
& -- & -- & -- & -- & -- & -- \\
\midrule
\multicolumn{13}{l}{\textit{LLM-Based}} \\
LFP~\cite{liu2023language}
& 30.55 & 59.59 & 76.86 & 15.97 & 50.70 & 75.30
& 30.64 & 54.72 & 76.86 & 15.97 & 50.70 & 75.30 \\
VidAssist (zero-shot)~\cite{islam2024propose}
& 14.60 & 52.60 & 68.38 & 9.89 & 40.85 & 70.35
& 18.44 & 50.63 & 75.64 & 9.07 & 42.72 & 80.83 \\
VidAssist~\cite{islam2024propose}
& 28.85 & 58.12 & 75.36 & 15.45 & 51.51 & 72.61
& 29.20 & 54.76 & 78.02 & 20.78 & 49.07 & 78.93 \\
SCHEMA~\cite{niu2024schema}
& 38.93 & 63.80 & 79.82 & 24.50 & 58.48 & 76.48
& 32.09 & 49.84 & 83.83 & 22.02 & 45.33 & 83.47 \\
\midrule
\multicolumn{13}{l}{\textit{Generative (World) Models}} \\
DDN~\cite{chang2020procedure}
& 12.18 & 31.29 & 47.48 & 5.97 & 27.10 & 48.46
& 13.90 & 20.19 & 64.78 & 11.13 & 17.71 & 68.06 \\
Int-MGAIL~\cite{bi2021procedure}
& 17.03 & 44.66 & 58.08 & 9.47 & 37.16 & 57.24
& -- & -- & -- & -- & -- & -- \\
Ext-MGAIL~\cite{bi2021procedure}
& 21.27 & 49.46 & 61.70 & 16.41 & 43.05 & 60.93
& -- & -- & -- & -- & -- & -- \\
P$^3$IV~\cite{zhao2022p3iv}
& 23.34 & 49.96 & 73.89 & 13.40 & 44.16 & 70.01
& 15.40 & 21.67 & 76.31 & 11.32 & 18.85 & 70.53 \\
PDPP~\cite{wang2023pdpp}
& 37.20 & 64.67 & 66.57 & 21.48 & 57.82 & 65.13
& 21.33 & 45.62 & 51.82 & 14.41 & 44.10 & 51.39 \\
KEPP~\cite{nagasinghe2024not}
& 38.12 & 64.74 & 67.15 & 24.15 & 59.05 & 66.64
& 20.25 & 39.87 & 51.72 & 15.63 & 39.53 & 53.27 \\
ActionDiffusion~\cite{shi2025actiondiffusion}
& 37.79 & 65.38 & 67.45 & 22.43 & 59.42 & 66.04
& 24.00 & 45.42 & 54.29 & 18.04 & 44.54 & 56.23 \\
MTID~\cite{zhou2025masked}
& 40.45 & 67.19 & 69.17 & 24.76 & 60.69 & 67.67
& 30.44 & 51.70 & 59.74 & 22.74 & 49.90 & 61.25 \\
\midrule
\multicolumn{13}{l}{\textit{Predictive (World) Models}} \\
WLTDO~\cite{ehsani2018let}
& 1.87 & 21.64 & 31.70 & 0.77 & 17.92 & 26.43
& -- & -- & -- & -- & -- & -- \\
UAAA~\cite{abu2019uncertainty}
& 2.15 & 20.21 & 30.87 & 0.98 & 19.86 & 27.09
& -- & -- & -- & -- & -- & -- \\
UPN~\cite{srinivas2018universal}
& 2.89 & 24.39 & 31.56 & 1.19 & 21.59 & 27.85
& -- & -- & -- & -- & -- & -- \\
PlaTe~\cite{sun2022plate}
& 16.00 & 36.17 & 65.91 & 14.00 & 35.29 & 55.36
& -- & -- & -- & -- & -- & -- \\
E3P~\cite{wang2023event}
& 26.40 & 53.02 & 74.05 & 16.49 & 48.00 & 70.16
& 19.57 & 31.42 & 84.95 & 13.59 & 26.72 & 84.72 \\
V-JEPA 2 ViT-L~\cite{assran2025v}
& 43.33 & 68.63 & 67.84 & 27.53 & 63.80 & 65.45
& 32.10 & 54.25 & 85.18 & 20.86 & 52.61 & 84.33 \\
V-JEPA 2 ViT-H~\cite{assran2025v}
& 44.07 & 70.18 & 68.32 & 28.75 & 64.71 & 66.82
& 32.76 & 55.37 & 86.57 & 22.60 & 53.19 & 85.74 \\
V-JEPA 2 ViT-g~\cite{assran2025v}
& 44.84 & 71.62 & 68.87 & 30.03 & 65.04 & 67.93
& 33.42 & 56.29 & 87.31 & 23.04 & 54.47 & 86.13 \\
V-JEPA 2 ViT-g$_{384}$~\cite{assran2025v}
& 45.58 & 72.74 & 69.42 & 31.36 & 65.45 & 69.21
& 34.08 & 57.20 & 89.53 & 23.43 & 55.58 & 86.57 \\
\rowcolor{cyan!20}GeoWorld ViT-L (Ours)
& 43.89 & 68.96 & 82.93 & 27.64 & 64.35 & 79.43
& 33.42 & 57.26 & 88.03 & 24.96 & 52.92 & 85.26 \\
\rowcolor{cyan!20}GeoWorld ViT-H (Ours)
& 45.33 & 70.84 & 84.70 & 29.19 & 65.47 & 80.16
& 34.08 & 58.70 & 88.42 & 26.24 & 53.66 & 87.17 \\
\rowcolor{cyan!20}GeoWorld ViT-g (Ours)
& 46.25 & 71.95 & 85.44 & 30.63 & 66.02 & 81.82
& 34.41 & 60.47 & 89.00 & 27.46 & 54.55 & 88.20 \\
\rowcolor{cyan!20} \textbf{GeoWorld ViT-g$_{384}$ (Ours)}
& \textbf{47.47} & \textbf{73.69} & \textbf{86.55} & \textbf{31.48} & \textbf{67.30} & \textbf{82.48}
& \textbf{34.85} & \textbf{61.86} & \textbf{89.88} & \textbf{27.79} & \textbf{55.97} & \textbf{88.61} \\
\bottomrule
\end{tabular}
}
\vspace{-2em}
\end{table*}

\subsection{Implementation Details}
\label{sec:implementation}

For a fair comparison, both the V-JEPA 2~\cite{assran2025v} baseline and our GeoWorld adopt frozen encoders pretrained on VideoMix22M. 
The exponential map $\exp_{0}(\cdot)$ is implemented and trained as a differentiable hyperbolic projection layer, where the curvature $c$ is treated as a learnable parameter~\cite{chami2019hyperbolic}.
The predictor network \(P_{\phi}(\cdot)\) is a \(\sim\!300\)M-parameter transformer with 24 layers, 16 heads, a 1024-dimensional hidden size, and GELU activations.

We conduct a two-stage training procedure for both V-JEPA~2 and our GeoWorld, consisting of supervised post-training followed by geometric reinforcement learning.

In the supervised post-training stage, both V-JEPA~2 and GeoWorld are trained with the AdamW optimizer~\cite{loshchilov2017decoupled} using a warmup–constant–decay learning rate schedule and a constant weight decay of 0.04. We linearly warm up the learning rate from \(7.5 \times 10^{-5}\) to \(4.25 \times 10^{-4}\) over 4500 iterations, hold it constant for 85{,}500 iterations, and then decay it to 0 over the final 4500 iterations, with a batch size of 256.

For geometric reinforcement learning, we keep the same AdamW optimizer and weight decay as in the supervised post-training stage, but adopt a smaller learning rate and a shorter schedule due to the higher variance of the RL objective. Specifically, we linearly warm up the learning rate from \(5.0 \times 10^{-5}\) to \(2.0 \times 10^{-4}\) over 2{,}000 iterations, hold it constant for 18{,}000 iterations, and then linearly decay it to 0 over the final 5{,}000 iterations, with a batch size of 128. Unless otherwise specified, we set the discount factor to \(\gamma = 0.99\) and the triangle-inequality regularization weight to \(\beta = 0.1\).

For energy-based planning with CEM~\cite{de2005tutorial}, we adopt a sample size of \(N = 800\), an elite set size of \(K = 80\), and \(I = 10\) refinement iterations.

The entire training is conducted on 4 nodes, each equipped with 8 NVIDIA H100 GPUs, 48-core Intel Xeon Platinum 8469C CPUs, and 230\,GB of RAM. We use only a single H100 GPU for inference.

\begin{table*}[t]
\centering
\small
\caption{\textbf{Goal-conditioned visual planning with videos on CrossTask~\cite{zhukov2019cross} and COIN~\cite{tang2019coin} datasets.}
We evaluate multi-step planning over a horizon \(T\) under the \textit{visual planning with videos}~\cite{ren2025videoworld} setup, where both observations and goals are specified as video clips.}
\label{tab:main_video}
\resizebox{\textwidth}{!}{
\begin{tabular}{lcccccc ccccccc}
\toprule
\multirow{3}{*}{Method}
& \multicolumn{6}{c}{CrossTask Dataset~\cite{zhukov2019cross}}
& \multicolumn{6}{c}{COIN Dataset~\cite{tang2019coin}}
\\
\cmidrule(lr){2-7}\cmidrule(lr){8-13}
& \multicolumn{3}{c}{T=3}
& \multicolumn{3}{c}{T=4}
& \multicolumn{3}{c}{T=3}
& \multicolumn{3}{c}{T=4}
\\
\cmidrule(lr){2-4}\cmidrule(lr){5-7}
\cmidrule(lr){8-10}\cmidrule(lr){11-13}
& SR & mAcc & mIoU
& SR & mAcc & mIoU
& SR & mAcc & mIoU
& SR & mAcc & mIoU
\\
\midrule
\multicolumn{13}{l}{\textit{LLM-Based}} \\
InternVL3.5-241B~\cite{wang2025internvl3_5}
& 44.03 & 70.01 & 84.41 & 27.65 & 63.54 & 80.13
& 36.54 & 57.22 & 89.02 & 25.46 & 55.30 & 88.22 \\
Qwen3-VL-Max~\cite{yang2025qwen3}
& 45.47 & 70.93 & 86.18 & 28.76 & 62.91 & 81.51
& 37.56 & 57.80 & 90.46 & 26.17 & 57.13 & 87.56 \\
Gemini 2.5 Pro~\cite{comanici2025gemini}
& 48.91 & 73.82 & 90.30 & 31.53 & 60.58 & 84.56
& 42.07 & 61.02 & 92.94 & 30.20 & 60.13 & 84.82 \\
GPT-5~\cite{openai2025gpt5systemcard}
& 50.03 & 72.38 & 91.18 & 30.20 & 64.48 & 82.15
& 43.84 & 64.67 & 91.12 & 32.64 & 56.84 & 86.38 \\
\midrule
\multicolumn{13}{l}{\textit{Generative (World) Models}} \\
VideoWorld~\cite{ren2025videoworld}
& 41.59 & 66.11 & 82.64 & 25.50 & 60.26 & 76.85
& 34.88 & 54.71 & 85.58 & 23.74 & 51.27 & 85.33 \\
\midrule
\multicolumn{13}{l}{\textit{Predictive (World) Models}} \\
V-JEPA 2 ViT-L~\cite{assran2025v}
& 43.36 & 69.55 & 84.75 & 28.86 & 64.34 & 78.40
& 36.10 & 56.70 & 87.02 & 25.29 & 53.30 & 86.21 \\
V-JEPA 2 ViT-H~\cite{assran2025v}
& 46.02 & 71.98 & 87.29 & 32.18 & 67.23 & 80.84
& 39.42 & 59.42 & 89.44 & 27.38 & 56.07 & 89.20 \\
V-JEPA 2 ViT-g~\cite{assran2025v}
& 48.13 & 73.42 & 89.62 & 33.46 & 69.26 & 82.61
& 40.97 & 61.86 & 90.77 & 29.60 & 57.73 & 90.23 \\
V-JEPA 2 ViT-g$_{384}$~\cite{assran2025v}
& 50.16 & 74.86 & 91.73 & 35.01 & 70.24 & 85.05
& 42.74 & 64.08 & 91.88 & 31.63 & 59.28 & 90.51 \\
\rowcolor{cyan!20}GeoWorld ViT-L (Ours)
& 44.80 & 70.54 & 86.30 & 30.63 & 65.46 & 79.73
& 37.76 & 58.14 & 88.00 & 26.40 & 54.52 & 86.98 \\
\rowcolor{cyan!20}GeoWorld ViT-H (Ours)
& 47.79 & 74.42 & 88.84 & 34.51 & 68.89 & 82.95
& 40.40 & 60.97 & 91.66 & 28.82 & 58.10 & 90.48 \\
\rowcolor{cyan!20}GeoWorld ViT-g (Ours)
& 49.23 & 76.64 & 90.61 & 35.49 & 71.00 & 84.50
& 42.84 & 62.63 & 93.69 & 29.93 & 60.65 & 91.81 \\
\rowcolor{cyan!20}GeoWorld ViT-g$_{384}$ (Ours)
& \textbf{51.71} & \textbf{77.30} & \textbf{92.95} & \textbf{37.04} & \textbf{71.35} & \textbf{87.04}
& \textbf{45.29} & \textbf{65.52} & \textbf{93.91} & \textbf{33.29} & \textbf{61.56} & \textbf{91.84} \\
\bottomrule
\end{tabular}
}
\vspace{-2em}
\end{table*}

\begin{table}[t]
\centering
\small
\caption{\textbf{Long horzion planning} on CrossTask~\cite{zhukov2019cross}.}
\label{tab:long}
\resizebox{\linewidth}{!}{
\begin{tabular}{lcccc}
\toprule
\multirow{2}{*}{Method}
& \multicolumn{4}{c}{Successful Rate (SR, \%)} \\
\cmidrule(lr){2-5}
& T=3 & T=4 & T=5 & T=6 \\
\midrule
\multicolumn{5}{c}{\textit{Procedural Planning (PP)}}\\
\midrule
Random~\cite{chang2020procedure}
& 0.01 & 0.01 & 0.01 & 0.01 \\
Retrieval-Based~\cite{zhou2025masked}
& 8.05 & 3.95 & 2.40 & 1.10 \\
DDN~\cite{chang2020procedure}
& 12.18 & 5.97 & 3.10 & 1.20 \\
P$^3$IV~\cite{zhao2022p3iv}
& 23.34 & 13.40 & 7.21 & 4.40 \\
E3P~\cite{wang2023event}
& 26.40 & 16.49 & 8.96 & 5.76 \\
PDPP~\cite{wang2023pdpp}
& 37.20 & 21.48 & 13.45 & 8.41 \\
KEPP~\cite{nagasinghe2024not}
& 38.12 & 24.15 & 14.20 & 9.27 \\
SCHEMA~\cite{niu2024schema}
& 38.93 & 24.50 & 14.75 & 10.53 \\
MTID~\cite{zhou2025masked}
& 40.45 & 24.76 & 15.26 & 10.30 \\
V-JEPA 2 ViT-L~\cite{assran2025v}
& 43.33 & 27.53 & 16.94 & 11.55 \\
\midrule
\textbf{GeoWorld ViT-L (Ours)}
& \textbf{43.89} & \textbf{27.64} & \textbf{17.38} & \textbf{12.37} \\
\midrule
\multicolumn{5}{c}{\textit{Visual Planning with Videos}}\\
\midrule
VideoWorld~\cite{ren2025videoworld}
& 41.59 & 25.50 & 15.36 & 10.97 \\
InternVL3.5-241B~\cite{wang2025internvl3_5}
& 44.03 & 27.65 & 17.31 & 12.44 \\
Qwen3-VL-Max~\cite{yang2025qwen3}
& 45.47 & 28.76 & 17.95 & 13.20 \\
Gemini 2.5 Pro~\cite{comanici2025gemini}
& 48.91 & 31.53 & 20.08 & 15.93 \\
GPT-5~\cite{openai2025gpt5systemcard}
& 50.03 & 30.20 & 21.46 & 16.07 \\
V-JEPA 2 ViT-g$_{384}$~\cite{assran2025v}
& 50.16 & 35.01 & 23.17 & 16.88 \\
\midrule
\textbf{GeoWorld ViT-g$_{384}$ (Ours)}
& \textbf{51.71} & \textbf{37.04} & \textbf{24.83} & \textbf{18.26} \\
\bottomrule
\end{tabular}
}
\vspace{-2em}
\end{table}

\subsection{Main Results}

As shown in Table~\ref{tab:main_pp} and~\ref{tab:main_video}, GeoWorld consistently improves multi-step goal-conditioned visual planning across both CrossTask and COIN. Under the procedural planning setup, GeoWorld yields notable gains over prior predictive world models, especially in long-horizon settings, achieving higher SR, mAcc, and mIoU for both \(T{=}3\) and \(T{=}4\). In the video-based planning setup, GeoWorld continues to outperform V-JEPA~2 across all model scales, with the ViT-g$_{384}$ variant achieving the best overall results and surpassing strong LLM-based planners. These improvements highlight the effectiveness of geometry-aware latent dynamics and geometric reinforcement learning in enhancing long-horizon stability and planning accuracy.

\textbf{For ablation study, please refer to \textit{Appendix}~\ref{sec:ablation}.}

\subsection{Long-Horizon Planning}
\label{sec:long}

Table~\ref{tab:long} highlights GeoWorld’s strength in long-horizon planning. As the horizon increases from $T=3$ to $T=6$, the performance of existing predictive and generative world models consistently degrades due to accumulated geometric drift in Euclidean latent space. In contrast, GeoWorld maintains higher stability and achieves the best Success Rate across all horizons. 

\section{Conclusion}

We introduced \textbf{GeoWorld}, a geometric world model designed to improve long-horizon visual planning by preserving geometric structure and hierarchical relations in latent space. Through \textit{Hyperbolic JEPA}, GeoWorld maps Euclidean latent representations onto a hyperbolic manifold, enabling geodesic-aware latent dynamics that produce a more structured and physically meaningful energy landscape. Building on this representation, \textit{Geometric Reinforcement Learning} refines the predictor via hyperbolic energy optimization and triangle-inequality regularization, yielding geodesic-consistent rollouts and reducing error accumulation across extended horizons. Extensive experiments on CrossTask and COIN demonstrate that GeoWorld consistently improves long-horizon performance over strong predictive world models such as V-JEPA~2, achieving higher success rates across \(T=3\) to \(T=6\) planning. These results highlight the importance of incorporating geometric principles into predictive world models and reinforce the value of geometry-aware reinforcement learning for stable and effective multi-step planning.

\clearpage
\clearpage
{
    \small
    \bibliographystyle{ieeenat_fullname}
    \bibliography{main}
}

\clearpage
\setcounter{page}{1}
\setcounter{section}{0}
\setcounter{table}{0}
\setcounter{figure}{0}
\maketitlesupplementary

\section{Preliminaries}
\label{sec:prelim}

\subsection{Energy-Based World Models}

Energy-Based World Models (EBWM) \cite{lecun2022path,assran2023self,bardes2024revisiting,assran2025v} are derived from Energy-Based Models (EBM) \cite{boney2020regularizing,du2022learning}, which define a scalar energy function $F(x, y)$ that measures how compatible two variables are, such as a current world state $s_x$ and a possible future state $s_y$. A low energy value corresponds to a plausible scenario, while a high energy value indicates an implausible one. Instead of predicting a single future, the energy landscape implicitly represents all plausible futures as valleys of low energy. Because energy replaces probability, the model can naturally handle multi-modal or uncertain worlds without the need for explicit sampling or normalization. Reasoning and planning are therefore formulated as energy minimization, where the goal is to find the configuration (actions, latents, or next states) that minimizes the expected energy:
\begin{equation}
\text{Plan} = \arg\min_{\text{actions}} F(s_t, s_{t+1:T}).
\end{equation}

\subsection{Hierarchical Planning}
\label{sec:hier}

Hierarchical planning often consists of two levels: high-level planning computes trajectories in abstract latent space that minimize energy (i.e., the most plausible and least costly transitions), while lower levels fill in the physical details \cite{lecun2022path}. 

The world model learns hierarchical latent abstractions, where low-level modules predict short-term fine details and higher-level modules capture long-term abstract dynamics.

Specifically, planning becomes energy minimization in latent space. High-level modules operate on abstract latent states $s^{(2)}$ that evolve slowly, where planning corresponds to finding a geodesic of minimum energy between the abstract states $s_A^{(2)}$ and $s_B^{(2)}$, yielding a coarse trajectory that serves as the overall plan. Lower levels then refine this trajectory into fine-grained predictions $s_t^{(1)}$, minimizing sub-energies conditioned on the higher-level plan. As a result, the overall behavior emerges through hierarchical energy descent, where each layer enforces consistency between its predictions and the layer above. In other words, high-level planning computes trajectories in abstract latent space that minimize energy (i.e., the most plausible and least costly transitions), while lower levels fill in the physical details.

Formally, for a hierarchical world model $F^{(L)}$, the optimal plan is defined as:
\begin{equation}
\text{Optimal Plan:} \quad 
\min_{\{a_t, z_t\}} 
\sum_{l=1}^{L} 
\sum_{t} 
F^{(l)}\big(s_t^{(l)}, s_{t+1}^{(l)}, z_t^{(l)}\big).
\end{equation}

Each $F^{(l)}$ expresses the energy cost of moving between abstract states at level $l$, and gradients through this hierarchy yield a coherent plan across scales.

\subsection{Joint-Embedding Predictive Architecture}
\label{sec:jepa}

Joint-Embedding Predictive Architectures (JEPA)~\cite{lecun2022path} learn a predictive world model directly in latent space rather than generating pixels. A JEPA encodes observations into a compact representation space and predicts future latent states by minimizing an energy or similarity objective between encoded targets and predicted embeddings. This joint-embedding formulation bypasses the need for autoregressive pixel generation, which is computationally expensive and prone to error accumulation over long horizons. Learning in latent space instead focuses the model on high-level structure, semantics, and temporal dependencies rather than low-level appearance details, enabling more stable and efficient multi-step prediction. By operating on representations rather than images, JEPA captures the underlying dynamics of the environment while avoiding the challenges of modeling raw pixel distributions.

\subsubsection{JEPA} 

We define the current observation $x$ and target $y$. $E_\theta(\cdot)$ denotes the observation encoder that maps raw visual inputs into the latent representation space, $\bar{E}_\theta(\cdot)$ denotes the target encoder with exponential moving average (EMA) weights, $s_x$ and $s_y$ are the latent representation of the current and target states obtained from the encoders, and $z$ is a latent variable capturing uncertainty. $P_\phi(\cdot)$ denotes the predictor, 
and $\theta$ and $\phi$ denote the parameters (weights) of the encoder and predictor networks, respectively.

For a unified one-step observation–target formulation, we first encode the current observation:

\begin{equation}
s_x = E_\theta(x)
\end{equation}

Then we perform latent prediction, which takes the current state $s_x$ and the uncertainty $z$ as input and predicts the latent representation of the target state $\hat{s}_y$:

\begin{equation}
\hat{s}_y = P_\phi(s_x, z)
\end{equation}

Similarly, we encode the target representation:

\begin{equation}
s_y = \bar{E}_\theta(y)
\end{equation}

For planning, the training objective becomes an energy minimization in latent space, which involves finding a geodesic of minimum energy between $s_x$ and $s_y$ to obtain the optimal plan.

\begin{equation}
\min_{z} \; C(s_x, s_y, z),
\quad \text{where } 
s_x = E_\theta(x), \; s_y = \bar{E}_\theta(y).
\end{equation}

Here, $C$ represents the energy cost of moving between the abstract states $s_x$ and $s_y$. It learns two encoders (for past and future) and a predictor that maps $s_x$ to $s_y$, where the energy is defined as the representation mismatch between the predicted and true embeddings. 
The JEPA is non-generative and is trained non-contrastively to ensure that the embeddings remain both informative and predictable~\cite{bardes2022vicreg}. 
As a result, it forms a predictive world model that learns latent abstractions.

\paragraph{Hierarchical-JEPA}
If we extend to hierarchical planning, we can develop a model that learns hierarchical latent abstractions (Hierarchical-JEPA), in which low-level modules predict short-term fine details, while higher-level modules capture long-term abstract dynamics. As an example, consider a two-level model where high-level JEPA layers operate on abstract latent states $s^{(2)}$ that evolve slowly.
Planning in this context corresponds to finding a geodesic of minimum energy between the abstract states $s_x^{(2)}$ and $s_y^{(2)}$, yielding a coarse trajectory, i.e., the plan.
Meanwhile, lower levels refine this trajectory into fine-grained predictions $s_t^{(1)}$, minimizing sub-energies conditioned on the higher-level plan. 
Overall behavior emerges through hierarchical energy descent, where each layer enforces consistency between its predictions and the layer above.

In other words, high-level planning computes trajectories in abstract latent space that minimize energy (i.e., most plausible and least costly transitions), while lower levels fill in the physical details.

Formally, the training objective of a Hierarchical-JEPA is given by:

\begin{equation}
\min_{\{z_t\}} 
\sum_{l=1}^{L} 
\sum_{t} 
C^{(l)}\big(s_t^{(l)}, s_{t+1}^{(l)}, z_t^{(l)}\big),
\end{equation}

where each $C^{(l)}$ represents the energy cost of moving between abstract states at level $l$, and gradients through this hierarchy yield a coherent plan across scales.

\subsubsection{I-JEPA}

Image-JEPA (I-JEPA)~\cite{assran2023self} extends JEPA to learn semantic image representations by predicting latent features of masked image regions from visible context patches. 

In I-JEPA, an image $y$ is divided into non-overlapping patches, from which a single large block is sampled as the context and several smaller blocks are sampled as targets. 
The context block $x$ is fed to the context encoder $E_\theta(\cdot)$ to obtain patch-level latent representations $s_x$, while the target blocks are processed by a target encoder with EMA weights, $\bar{E}_\theta(\cdot)$, to produce target embeddings $s_y$. 
The predictor network $P_\phi(\cdot)$ takes the context representation $s_x$ along with positional mask tokens $\{m_j\}_{j \in B_i}$ indicating the spatial locations of each target block, and predicts the corresponding feature vectors $\hat{s}_y = P_\phi(s_x, \{m_j\}_{j \in B_i})$ for those regions. 

The training objective minimizes the average squared distance between the predicted and target representations of the target blocks, averaged over all sampled blocks and training samples in the dataset $\mathcal{D}$:

\begin{multline}
\min_{\theta, \phi} \;
\mathbb{E}_{(x, y) \sim \mathcal{D}}
\Bigg[
\frac{1}{M} \sum_{i=1}^{M} \sum_{j \in B_i}
\Big\|
P_\phi\big(E_\theta(x), \{m_j\}_{j \in B_i}\big)
\\
- \bar{E}_\theta(y)_j
\Big\|_2^2
\Bigg].
\end{multline}

Here, $M$ denotes the number of target blocks, $\mathbb{E}[\cdot]$ denotes the expectation (average) over all training samples $(x, y)$, and the loss measures the representation-level prediction error rather than pixel-level reconstruction, allowing I-JEPA to learn highly semantic, non-generative representations.

\subsubsection{V-JEPA}
\label{sec:v-jepa}

Video-JEPA (V-JEPA)~\cite{bardes2024revisiting} and V-JEPA 2~\cite{assran2025v} extend JEPA to learn spatio-temporal video representations by predicting masked tubelet features from visible context regions. 

In V-JEPA and V-JEPA 2, a video clip $y$ is tokenized into spatial–temporal patches (tubelets), and a subset of these patches is masked (replaced with mask tokens). 
The remaining unmasked patches form the context view, while the masked patches form the target view. 
The observation (context) encoder $E_\theta(\cdot)$ processes the masked version $x$ (containing both visible and mask tokens) and produces latent embeddings for all positions, including the masked ones. 
However, the predictor $P_\phi(\cdot)$ focuses only on the masked positions. 
A target encoder with EMA weights, $\bar{E}_\theta(\cdot)$, computes the target embeddings $s_y$ of the full (unmasked) input. 
The predictor takes the context representation $s_x = E_\theta(x)$ along with the mask token indicators $\Delta_y$ and predicts the feature vectors $\hat{s}_y$ for each masked patch position.

The training objective is to minimize the $L_1$ distance between the predicted representations of the masked regions and the target representations from the target encoder, averaged over the dataset $\mathcal{D}$:

\begin{equation}
\min_{\theta, \phi} \; \mathbb{E}_{(x, \Delta_y, y) \sim \mathcal{D}}
\left\|
P_\phi\big(E_\theta(x), \Delta_y\big) - \bar{E}_\theta(y)
\right\|_1
\end{equation}

where $\mathbb{E}[\cdot]$ denotes the expectation (average) over all training samples $(x, \Delta_y, y)$.

That is,

\begin{equation}
\min_{\theta, \phi} \; \mathbb{E}_{(s_x, \Delta_y, s_y) \sim \mathcal{D}}
\left[
\left\|
\hat{s}_y - s_y
\right\|_1
\right],
~\text{where } 
\hat{s}_y = P_\phi(s_x, \Delta_y).
\end{equation}

Hence the loss function $\mathcal{L}$ measuring the $L_1$ distance between the $\hat{s}_y$ and $s_y$:

\begin{equation}
\mathcal{L}(\theta, \phi) = 
\left\|
P_\phi\big(E_\theta(x), \Delta_y) - \bar{E}_\theta(y)
\right\|_1
\end{equation}

\paragraph{V-JEPA 2-AC}
The action-conditioned variation (V-JEPA 2-AC) serves as a downstream extension of V-JEPA 2 that predicts future latent representations conditioned on agent actions. 

As a step-by-step (1-step prediction) formulation, V-JEPA 2-AC adapts the frozen encoder $E(\cdot)$ from V-JEPA 2, pretrained on unlabeled videos, to encode the current observation $x_t$ into the latent representation $s_t^{x}$ and the future target $x_{t+1}$ into $s_{t+1}^{x}$. 
The action-conditioned predictor $P_\phi(\cdot)$ takes the current latent $s_t^{x}$ and the action $a_t$ as input to predict the next-state latent representation $\hat{s}_{t+1}^{x}$. 
The $L_1$ loss in latent space then trains the predictor to align its predicted next-state representations with the encoded future representations:

\begin{align}
\mathcal{L}_{\text{AC}}(\phi) &= 
\left\|
P_\phi\big(E(x_t), a_t\big) - E(x_{t+1})
\right\|_1 \\
&= 
\left\|
\hat{s}_{t+1}^{x} - s_{t+1}^{x}
\right\|_1.
\end{align}

In the multi-step rollout setting, V-JEPA 2-AC extends the one-step formulation to predict a sequence of future latent representations over a time horizon $T$ conditioned on a sequence of actions. 

We randomly sample a mini-batch of 4-second video clips from the Droid dataset~\cite{khazatsky2024droid} and, for simplicity, discard any videos shorter than 4 seconds, leaving us with a smaller subset of the dataset comprising under 62 hours of video. The video clips are sampled with a resolution of $256 \times 256$ and a frame rate of 4 fps, yielding 16-frame clips $(x_t)_{t=1}^{16}$, where
each $x_t$ represents a single video frame. The robot’s end-effector state in each observation is denoted by the sequence $(s_t^{e})_{t=1}^{16}$, where $s_t^{e}$ is a real-valued 7D vector defined relative to the base of the robot. We construct a sequence of actions $(a_t)_{t=1}^{15}$ by computing the change in end-effector state between adjacent frames. We use the V-JEPA~2 encoder $E(\cdot)$ as an image encoder and encode each frame independently in a given clip to obtain a sequence of feature maps $(s_t^{x})_{t=1}^{16}$. The sequence of observed feature maps, end-effector states, and actions is temporally interleaved as $(s_t^{x}, s_t^{e}, a_t)_{t=1}^{15}$ and processed with the transformer predictor network $P_{\phi}(\cdot)$ to obtain a sequence of next-state representation predictions $(s_{t+1}^{x})_{t=1}^{15}$:

\begin{equation}
(\hat{s}_{t+1}^{x})_{t=1}^{15}
= 
P_{\phi}\!\big((s_t^{x}, s_t^{e}, a_t)_{t=1}^{15}\big).
\end{equation}

The teacher-forcing loss trains the predictor $P_{\phi}(\cdot)$ to accurately perform one-step future prediction by minimizing the $L_1$ distance between the predicted latent representation $\hat{s}_{t+1}^{x}$ and the encoded ground-truth latent $s_{t+1}^{x}$ at each time step $t$:

\begin{align}
\mathcal{L}_{\text{TF}}(\phi)
&= 
\frac{1}{T}\sum_{t=1}^{T}
\left\|
P_{\phi}(E(x_t), s_t^{e}, a_t) - E(x_{t+1})
\right\|_1 \\
&=
\frac{1}{T}\sum_{t=1}^{T}
\left\|
\hat{s}_{t+1}^{x} - s_{t+1}^{x}
\right\|_1,
~\text{where } T = 15.
\end{align}

The rollout loss involves feeding the predictor’s output back as input, allowing the model to be trained to predict several timesteps ahead. In this case, we design a two-step rollout loss to improve the model’s ability to perform autoregressive rollouts during inference. We can now denote the rollout loss as

\begin{align}
\mathcal{L}_{\text{rollout}}(\phi)
&=
\left\|
P_{\phi}(E(x_t), s_t^{e}, (a_t)_{t}^{t + n})
- E(x_{t + n + 1})
\right\|_1,\\
&=
\left\|
P_{\phi}(s_t^{x}, s_t^{e}, a_t, a_{t+1})
- s_{t+2}^{x})
\right\|_1,\\
&=
\left\|
\hat{s}_{t+2}^{x} - s_{t+2}^{x}
\right\|_1,
~\text{where } n = 1.
\end{align}

Hence, the total loss is

\begin{equation}
\mathcal{L}_{\text{AC}}(\phi)
=
\mathcal{L}_{\text{TF}}(\phi)
+
\mathcal{L}_{\text{rollout}}(\phi).
\end{equation}

\paragraph{Inference}

We can then perform energy-based planning after training, using the frozen encoder $E$ and predictor $P$. This predictor acts as the world model, capable of imagining how latent representations evolve when an action sequence is applied. At test time, no weights are trained, instead, we search for an action sequence that minimizes a goal-conditioned energy cost between the imagined future and the goal latent representation.

Given the current observation $x_1$, current end-effector state $s_1^{e}$, target goal image $x_{1+T}$, and planning horizon $T$, we encode the current and goal observations as:
\begin{equation}
s_1^{x} = E(x_1), \qquad s_{1+T}^{x} = E(x_{1+T}).
\end{equation}

We then define the $L_1$ energy cost function:
\begin{equation}
C((\hat{a}_t)_{t=1}^{T}; s_1^{e}, s_1^{x}, s_{1+T}^{x})
= \left\|
P((\hat{a}_t)_{t=1}^{T}; s_1^{e}, s_1^{x}) - s_{1+T}^{x}
\right\|_1.
\label{eq:energy_cost}
\end{equation}

The optimal action sequence is obtained by minimizing this cost:
\begin{equation}
(a_t^{*})_{t=1}^{T}
= \arg\min_{(\hat{a}_t)_{t=1}^{T}}
C((\hat{a}_t)_{t=1}^{T}; s_1^{e}, s_1^{x}, s_{1+T}^{x}).
\end{equation}

Thus, the predictor is used to imagine the future latent trajectory, and planning reduces to finding the action sequence that minimizes the latent L1 distance to the goal embedding.

V-JEPA 2-AC uses the Cross-Entropy Method (CEM)~\cite{de2005tutorial} with 800 samples and 10 iterations to efficiently minimize $C$ at each planning step, as shown in Algorithm \ref{alg:cem_planning}. It executes only the first action on the robot before re-planning, as in receding horizon control, and is tested only on a horizon of $T = 1$.

\begin{algorithm}[t]
\caption{Energy-Based Planning with Cross-Entropy Method (CEM)}
\label{alg:cem_planning}
\begin{algorithmic}[0]
\Require 
Predictor $P$ (world model), encoder $E(\cdot)$, planning horizon $T$, 
number of samples $N$, number of elites $K$, number of iterations $I$, 
initial mean $\mu_0$, and covariance $\Sigma_0$.
\Ensure 
Optimal action sequence $(a_t^{*})_{t=1}^{T}$.

\State \textbf{Input:} current observation $x_1$, end-effector state $s_1^{e}$, and goal image $x_{1+T}$.
\State Encode current and goal observations:
$s_1^{x} = E(x_1)$, $s_{1+T}^{x} = E(x_{1+T})$.
\vspace{2pt}

\For{$j = 1$ \textbf{to} $I$}
    \State \textbf{(1) Sample:} Draw $N$ candidate action sequences 
    $\{\hat{a}_{1:T}^{(n)}\}_{n=1}^{N}$ from 
    $\mathcal{N}(\mu_{j-1}, \Sigma_{j-1})$.
    \vspace{2pt}

    \State \textbf{(2) Evaluate:} For each candidate sequence, compute its energy cost:

    {\setlength{\abovedisplayskip}{-8pt}
    \setlength{\belowdisplayskip}{3pt}
    \[
    C^{(n)} = 
    \left\|
    P_{\phi}(\hat{a}_{1:T}^{(n)}; s_1^{e}, s_1^{x})
    - s_{1+T}^{x}
    \right\|_1.
    \]
    }

    \State \textbf{(3) Select elites:} Sort $\{C^{(n)}\}$ in ascending order
    and select the top $K$ sequences with the lowest cost to form the elite set 
    $\mathcal{E}_j = \{\hat{a}_{1:T}^{(n)} \mid n \in \text{top-}K(C)\}$.
    These elites represent trajectories that drive the world model's imagined latent 
    state closest to the goal latent $s_{1+T}^{x}$.
    \vspace{2pt}

    \State \textbf{(4) Update distribution:} 
    Compute the new mean and covariance of the elite set:
    {
    \setlength{\abovedisplayskip}{3pt}
    \setlength{\belowdisplayskip}{3pt}
    \begin{align*}
    \mu_j &= \frac{1}{K} \sum_{(\hat{a}_t)_{t=1}^{T} \in \mathcal{E}_j} (\hat{a}_t)_{t=1}^{T}, \nonumber \\
    \Sigma_j &= \frac{1}{K} \sum_{(\hat{a}_t)_{t=1}^{T} \in \mathcal{E}_j}
    \big( (\hat{a}_t)_{t=1}^{T} - \mu_j \big) \big( (\hat{a}_t)_{t=1}^{T} - \mu_j \big)^{\top}.
    \end{align*}
    }
    The new distribution $\mathcal{N}(\mu_j, \Sigma_j)$ is now centered around promising low-energy action sequences.
    \vspace{2pt}

    \State \textbf{(5) Repeat:} Continue iterating Steps~3–4 for $I$ iterations.
    As the process proceeds, the sampling distribution progressively concentrates
    around action sequences that minimize the latent-space cost $C_{\text{L1}}$.
\EndFor

\State \textbf{(6) Execute:} 
Select the action sequence corresponding to the lowest final cost:
{
\setlength{\abovedisplayskip}{3pt}
\setlength{\belowdisplayskip}{3pt}
\[
(a_t^{*})_{t=1}^{T}
= \arg\min_{(\hat{a}_t)_{t=1}^{T}}
C((\hat{a}_t)_{t=1}^{T}; s_1^{e}, s_1^{x}, s_{1+T}^{x}).
\]
}
and execute only the first action $a_1^{*}$ on the robot.
\vspace{2pt}

\State \textbf{(7) Re-plan:} 
Observe the next frame $x_2$, re-encode $s_2^{x} = E(x_2)$, 
and repeat the process from Step~1 (receding horizon control).

\end{algorithmic}
\end{algorithm}

In other words, during inference, the predictor $P$ serves as a world model for \textit{energy-based planning}, where at each step the CEM searches for an action sequence that minimizes the latent-space cost $C$ between the imagined future and the goal representation, executing only the first action $a_1^*$ before re-planning.

\subsection{World Modeling Paradigms}

There are two typical approaches for goal-conditional world modeling: generative world models \cite{mendonca2023structured,ren2025videoworld,singh2025hand} and predictive world models \cite{lecun2022path,assran2023self,garrido2024learning,bardes2024revisiting,assran2025v}.

\paragraph{Generative world models.}

Generative world models typically build upon autoregressive (AR) transformers or semi-AR (autoregressive diffusion) models $\rho$ that observe the visual context $x_t$ and autoregressively predict the latent $z_t$ and the next frame $\hat{x}_{t+1}$:

\begin{equation}
(z_t, \hat{x}_{t+1}) = \rho(x_t).
\end{equation}

They often rely on an inverse dynamics model (IDM) $\pi$ \cite{spong2002adaptive}, trained separately, which maps the pair $(x_t, z_t, \hat{x}_{t+1})$ to an explicit action \cite{yang2019imitation,brandfonbrener2023inverse,tian2025predictive}:

\begin{equation}
a_t = \pi(x_t, z_t, \hat{x}_{t+1}).
\end{equation}

In other words, it’s a one-step inverse mapping of the environment’s dynamics. Hence, it can only predict one step at a time, because it does not know the full trajectory structure or the energy landscape over multiple steps.

Moreover, it must explicitly generate or reconstruct the next frame (or latent visual tokens that decode into pixels), and perform planning by predicting how the world will look after an implicit action, which connects to the challenge mentioned in~\ref{sec:jepa}.

\paragraph{Predictive world models}

Predictive world models are not generative, they do not model pixel distributions. Instead, they learn an energy landscape in latent space that measures compatibility between current and target states.

However, predictive world models require an explicit goal observation $x_{t+1}$ to compute their goal-conditioned energy, as shown in Equation \ref{eq:energy_cost}, and minimize the energy cost with the CEM for planning. 

This design trade-off is intentional, not accidental. As mentioned in~\ref{sec:jepa}, we need to avoid pixel prediction during planning since pixels are noisy, unimportant, and computationally expensive~\cite{lecun2022path}. Therefore, predictive world models are not intended to be self-contained simulators. Moreover, unlike the IDM, which predicts only a single action given consecutive states, CEM performs multi-step trajectory optimization by searching over candidate action sequences to minimize the latent-space energy cost, enabling long-horizon planning rather than one-step reactive control.

\subsection{Hyperbolic Learning}

Hyperbolic space, denoted as $\mathbb{H}^n$, is a negatively curved Riemannian manifold characterized by exponential volume growth and a saddle-shaped geometry \cite{bridson2013metric}. Unlike Euclidean space, where parallel lines remain equidistant, lines in hyperbolic space diverge, and the volume expands exponentially with radius. This property makes hyperbolic geometry naturally suited for representing hierarchical or tree-like data structures \cite{nickel2017poincare,pal2024compositional}, such as hierarchical planning for world models, where the number of nodes grows exponentially with depth. In deep learning, hyperbolic space enables exponentially efficient representations of hierarchies by compressing large-scale differences while maintaining fine-grained local relationships. As a result, it has been widely applied in representation learning \cite{nickel2017poincare,yue2023hyperbolic,desai2023hyperbolic,pal2024compositional}, computer vision and graphics \cite{ge2023hyperbolic} that require modeling multi-level, non-Euclidean structures.

Hyperbolic space $\mathbb{H}^n$ is an abstract Riemannian manifold of constant negative curvature that does not depend on any coordinate system. In order to represent points in this curved space for computation, coordinate models are introduced to map $\mathbb{H}^n$ into Euclidean space while preserving its geometric structure. Two of the most common models are the Poincaré ball model $\mathbb{B}^n$ \cite{nickel2017poincare,yue2023hyperbolic,ge2023hyperbolic,cetin2023hyperbolic} and the Lorentz (or hyperboloid) model $\mathbb{L}^n$ \cite{desai2023hyperbolic,pal2024compositional}, both providing isometric representations of the same manifold but differing in their coordinate systems and numerical properties.

\subsubsection{Poincaré Ball Model.}  
\label{sec:poincare}
The Poincaré ball model $\mathbb{B}^n$ represents hyperbolic space as an open unit ball embedded in Euclidean space, defined as
\begin{equation}
\mathbb{B}^n = \{z \in \mathbb{R}^n : \|z\| < 1\}.
\end{equation}
It is endowed with a Riemannian metric that encodes constant negative curvature, ensuring that Euclidean distances are reweighted to reflect hyperbolic geometry. Each point $z$ lies strictly inside the ball, and distances are measured using the hyperbolic metric rather than Euclidean norms. This bounded representation makes the geometry intuitive and well-suited for visualization, as tree-like hierarchies naturally fit inside a finite domain where the boundary corresponds to infinite distance. 
It constrains embeddings to normalized radii, but note that in the Poincaré model “normalization” means keeping points \textit{inside} the unit ball, not unit-norm on a sphere (that corresponds to the hyperspherical case). 

\paragraph{Geodesics}
The geodesic, or the Poincaré-ball hyperbolic distance, between two points $u, v \in \mathbb{B}^n$ is a circular arc perpendicular to the boundary of the ball. 
Its length is given by the hyperbolic distance function~\cite{nickel2017poincare,ganea2018hyperbolic}
\begin{equation}
d_{\mathbb{H}}(u, v) = 
\operatorname{arcosh}\!\left(
1 + 2
\frac{\|u - v\|^2}
{(1 - \|u\|^2)(1 - \|v\|^2)}
\right).
\end{equation}
This metric measures the shortest path along the curved manifold rather than in Euclidean space, capturing the exponential growth of distances as points approach the boundary of the Poincaré ball.

\paragraph{Exponential Map}
In Riemannian geometry, the exponential map
\begin{equation}
    \exp_x : \mathbf{T}_x\mathbb{H}^n \rightarrow \mathbb{H}^n 
\end{equation}

takes a tangent vector $v \in \mathbf{T}_x\mathbb{H}^n$ (the tangent space at point $x$) and moves it along the geodesic starting from $x$ in direction $v$, traveling a distance equal to $\|v\|$ under the hyperbolic metric. In other words, $\exp_x(v)$ can be interpreted as starting at $x$ and walking along the manifold in the direction of $v$ for a distance $\|v\|$. This operation maps local Euclidean updates $v$ into global manifold coordinates, ensuring that updates remain consistent with the geometry of hyperbolic space. In hyperbolic space, the mapping from the tangent space to the manifold through the exponential map is one-to-one. Although there exist manifolds equipped with hyperbolic metrics where this mapping is not one-to-one~\cite{zhang2025flashmo}, the Poincaré model of hyperbolic space preserves this one-to-one correspondence, ensuring a well-defined relationship between the tangent space and the manifold.

The exponential map from the origin of the Poincaré ball, denoted as $\exp_0(v)$, maps Euclidean vectors directly into hyperbolic space, which is particularly useful for initialization. It is defined as
\begin{equation}
\exp_0(v) = \tanh(\|v\|)\frac{v}{\|v\|}.
\end{equation}
This formulation is simple because the tangent space at the origin aligns perfectly with the Euclidean space, making the mapping between Euclidean and hyperbolic representations straightforward.

The exponential map from a general point $x \in \mathbb{B}^n$ must account for the curvature around $x$. It moves the point $x$ along the geodesic in the direction of the tangent vector $v$. The key difference from the origin case is that $x$ is not the origin, so we need a way to “add” $v$ to $x$ under hyperbolic geometry. In Euclidean space, moving from a point $x$ by a vector $v$ is simply computed as $x + v$. However, in hyperbolic geometry, vector addition is replaced by \textit{Möbius addition}, denoted as $x \oplus v$. Therefore, the general exponential map is expressed as
\begin{equation}
\exp_x(v) = x \oplus \left( \tanh\!\left(\frac{\lambda_x \|v\|}{2}\right) \frac{v}{\|v\|} \right),
\label{eq:exp}
\end{equation}
where
\begin{equation}
\lambda_x = \frac{2}{1 - \|x\|^2}
\end{equation}
is the \textit{conformal factor} that rescales distances locally. This formulation ensures that the operation respects hyperbolic curvature instead of Euclidean linearity.

The Möbius addition of two vectors $x$ and $y$ is defined as
\begin{equation}
x \oplus y =
\frac{(1 + 2\langle x, y \rangle + \|y\|^2)x + (1 - \|x\|^2)y}
     {1 + 2\langle x, y \rangle + \|x\|^2 \|y\|^2}.
\label{eq:mobius_add}
\end{equation}
Substituting $y = \tanh\!\left(\frac{\lambda_x \|v\|}{2}\right)\frac{v}{\|v\|}$, we obtain an explicit expression for $\exp_x(v)$ in hyperbolic space

\begin{equation}
\exp_x(v) =
\frac{
\big(1 + 2\alpha \langle x, \tfrac{v}{\|v\|}\rangle + \alpha^2 \big)x
+ (1 - \|x\|^2)\alpha \tfrac{v}{\|v\|}}
{1 + 2\alpha \langle x, \tfrac{v}{\|v\|}\rangle + \alpha^2 \|x\|^2},
\end{equation}

where

\begin{equation}
\alpha = \tanh\!\left(\frac{\lambda_x \|v\|}{2}\right),
\quad
\lambda_x = \frac{2}{1 - \|x\|^2}.
\end{equation}
This formulation explicitly shows how the exponential map combines the curvature-adjusted scaling (via $\lambda_x$ and $\tanh$) with the non-linear composition of $x$ and $v$ under Möbius addition, ensuring consistency with hyperbolic geometry.

\paragraph{Logarithmic Map}
The logarithmic map serves as the inverse of the exponential map, mapping points from the manifold back to the tangent space at a given point $x \in \mathbb{B}^n$. Formally, it is defined as
\begin{equation}
\log_x : \mathbb{B}^n \rightarrow \mathbf{T}_x\mathbb{H}^n,
\end{equation}
which takes a point $y \in \mathbb{B}^n$ and returns a tangent vector $v \in \mathbf{T}_x\mathbb{H}^n$ that, when re-projected through the exponential map, satisfies $\exp_x(v) = y$. This operation locally linearizes the manifold around $x$, allowing differential computations such as gradient-based optimization to be performed in the tangent space.

The logarithmic map from the origin of the Poincaré ball, denoted as $\log_0(y)$, converts a point $y \in \mathbb{B}^n$ back to its Euclidean tangent vector and is defined as
\begin{equation}
\log_0(y) = \operatorname{arctanh}(\|y\|)\frac{y}{\|y\|}.
\end{equation}
This formulation is the inverse of the exponential map at the origin, satisfying $\exp_0(\log_0(y)) = y$, and is computationally simple since the tangent space at the origin coincides with $\mathbb{R}^n$.

For a general point $x \in \mathbb{B}^n$, the logarithmic map must account for the local curvature around $x$. It can be expressed using the inverse of the Möbius addition:
\begin{equation}
\log_x(y) = 
\frac{2}{\lambda_x}
\operatorname{arctanh}\!\big(\|{-x} \oplus y\|\big)
\frac{{-x} \oplus y}{\|{-x} \oplus y\|},
\label{eq:log}
\end{equation}
where
\begin{equation}
\lambda_x = \frac{2}{1 - \|x\|^2}
\end{equation}
is the same conformal factor as in the exponential map, and ${-x} \oplus y$ denotes Möbius addition with the inverse of $x$. 

Expanding the definition of the Möbius addition into the logarithmic map expression, we obtain an explicit formulation of $\log_x(y)$ in the Poincaré ball model.  
Substituting Eq.~\ref{eq:mobius_add}
into Eq.~\ref{eq:log}, we first compute the intermediate term ${-x} \oplus y$ as
\begin{equation}
{-x} \oplus y =
\frac{(1 - 2\langle x, y \rangle + \|y\|^2)(-x) + (1 - \|x\|^2)y}
     {1 - 2\langle x, y \rangle + \|x\|^2 \|y\|^2}.
\end{equation}
Then, the logarithmic map can be written explicitly as
\begin{equation}
\log_x(y)
= \frac{2}{\lambda_x}\,
\operatorname{arctanh}\!\left(\frac{\|N(x,y)\|}{D(x,y)}\right)\,
\frac{N(x,y)}{\|N(x,y)\|},
\end{equation}
where
\begin{align}
N(x,y) &= \big(1 - 2\langle x, y\rangle + \|y\|^2\big)(-x)
        + \big(1 - \|x\|^2\big)\,y,\\
D(x,y) &= 1 - 2\langle x, y\rangle + \|x\|^2\|y\|^2,
\quad
\lambda_x = \frac{2}{1-\|x\|^2}.
\end{align}

This expanded form explicitly expresses the logarithmic map in terms of $x$ and $y$, showing how the non-linear geometry of the Poincaré ball modifies vector displacement through the Möbius addition. It provides the tangent vector at $x$ that points toward $y$ with a magnitude corresponding to the hyperbolic distance between them.

This formulation ensures that $\log_x(y)$ returns a tangent vector at $x$ whose exponential map precisely recovers $y$, i.e., $\exp_x(\log_x(y)) = y$. Together, the exponential and logarithmic maps establish a smooth and invertible correspondence between the Euclidean tangent space and the curved manifold, enabling consistent optimization and representation learning in hyperbolic space.

\paragraph{Curvature}
The above formulation corresponds to the \textit{unit-curvature} case, where the curvature is fixed as $K = -1$. 
In general, hyperbolic space has a constant negative curvature usually written as $K = -c$, where $c > 0$. 
When $c = 1$, the space has curvature $-1$, which is the normalized convention adopted by most works. 
The general form instead keeps $c$ as a free curvature parameter, so the ball’s radius becomes $1 / \sqrt{c}$. 
This allows different degrees of curvature — flatter when $c \to 0$, and more curved when $c \to \infty$.

\begin{equation}
\mathbb{B}^n = \{z \in \mathbb{R}^n : c\|z\|^2 < 1\},
\end{equation}
where the radius is $1 / \sqrt{c}$ and curvature $K = -c$. 
The Riemannian metric is scaled by the \textit{conformal factor}
\begin{equation}
\lambda_x = \frac{2}{1 - c\|x\|^2},
\end{equation}
which defines the metric tensor as
\begin{equation}
\mathbf{G}_x = \lambda_x^2 \mathbf{I}.
\end{equation}

The general form of the Poincaré-ball hyperbolic distance under curvature $K = -c$ $(c > 0)$ between two points $u, v \in \mathbb{B}^n$ is given by~\cite{cetin2023hyperbolic}
\begin{equation}
d_{\mathbb{H}}(u, v) = 
\frac{1}{\sqrt{c}}
\operatorname{arcosh}\!\left(
1 + 2c
\frac{\|u - v\|^2}
{(1 - c\|u\|^2)(1 - c\|v\|^2)}
\right).
\label{eq:poincare_distance_c}
\end{equation}

The exponential map with the curvature $c$ is expressed as
\begin{equation}
\exp_x(v) = x \oplus_c \left( \tanh\!\left(\frac{\sqrt{c}\,\lambda_x \|v\|}{2}\right) 
\frac{v}{\sqrt{c}\,\|v\|} \right),
\end{equation}
where
\begin{equation}
\lambda_x = \frac{2}{1 - c\|x\|^2},
\end{equation}
and $\oplus_c$ denotes the Möbius addition under curvature $c$. 

The Möbius addition of two vectors $x$ and $y$ under curvature $c$ is defined as~\cite{ganea2018hyperbolic,cetin2023hyperbolic}
\begin{equation}
x \oplus_c y =
\frac{(1 + 2c\langle x, y \rangle + c\|y\|^2)x + (1 - c\|x\|^2)y}
     {1 + 2c\langle x, y \rangle + c^2\|x\|^2\|y\|^2}.
\label{eq:mobius_c}
\end{equation}

So the explicit expression for $\exp_x(v)$ in hyperbolic space with curvature $c$ is given by
\begin{equation}
\exp_x(v) =
\frac{
\big(1 + 2c\alpha \langle x, \tfrac{v}{\|v\|}\rangle + c\alpha^2 \|v\|^2 \big)x
+ (1 - c\|x\|^2)\alpha v}
{1 + 2c\alpha \langle x, \tfrac{v}{\|v\|}\rangle + c^2\alpha^2 \|x\|^2 \|v\|^2},
\label{eq:exp_c}
\end{equation}
where
\begin{equation}
\alpha = \tanh\!\left(\frac{\sqrt{c}\,\lambda_x \|v\|}{2}\right),
\quad
\lambda_x = \frac{2}{1 - c\|x\|^2}.
\end{equation}

For a general curvature $K = -c$ $(c > 0)$, the logarithmic map must account for the local curvature around $x$. It can be expressed using the inverse of the Möbius addition as
\begin{equation}
\log_x(y) = 
\frac{2}{\lambda_x \sqrt{c}}\,
\operatorname{arctanh}\!\big(\sqrt{c}\|{-x} \oplus_c y\|\big)
\frac{{-x} \oplus_c y}{\|{-x} \oplus_c y\|},
\label{eq:log_c}
\end{equation}
where
\begin{equation}
\lambda_x = \frac{2}{1 - c\|x\|^2}
\end{equation}
is the conformal factor, and ${-x} \oplus_c y$ denotes the Möbius addition under curvature $-c$.

Expanding the definition of Möbius addition into the logarithmic map expression, we obtain an explicit formulation of $\log_x(y)$ in the general Poincaré ball model with curvature $-c$. Substituting the $c$-dependent Möbius addition (Eq.~\ref{eq:mobius_c})
into Eq.~\ref{eq:log_c}, we first compute the intermediate term ${-x} \oplus_c y$ as
\begin{equation}
{-x} \oplus_c y =
\frac{(1 - 2c\langle x, y \rangle + c\|y\|^2)(-x) + (1 - c\|x\|^2)y}
     {1 - 2c\langle x, y \rangle + c^2\|x\|^2\|y\|^2}.
\end{equation}
Then, the logarithmic map can be written explicitly as
\begin{equation}
\log_x(y)
= \frac{2}{\lambda_x \sqrt{c}}\,
\operatorname{arctanh}\!\left(\sqrt{c}\frac{\|N_c(x,y)\|}{D_c(x,y)}\right)\,
\frac{N_c(x,y)}{\|N_c(x,y)\|},
\end{equation}
where
\begin{align}
N_c(x,y) &= \big(1 - 2c\langle x, y\rangle + c\|y\|^2\big)(-x)
        + \big(1 - c\|x\|^2\big)\,y,\\
D_c(x,y) &= 1 - 2c\langle x, y\rangle + c^2\|x\|^2\|y\|^2,
\quad
\lambda_x = \frac{2}{1 - c\|x\|^2}.
\end{align}

When $c = 1$, this expression reduces to the unit-curvature form of the logarithmic map given in Eq.~\ref{eq:log}.

\paragraph{Lorentz Model.}  
The Lorentz model $\mathbb{L}^n$ represents hyperbolic space as the upper sheet of a two-sheeted hyperboloid embedded in Minkowski space $\mathbb{R}^{n+1}$, defined as 
\begin{equation}
\mathbb{L}^n = \{p \in \mathbb{R}^{n+1} : \langle p, p \rangle_L = -1 / \kappa,\, p_0 > 0\},
\end{equation}
where $\langle p, q \rangle_L = -p_0 q_0 + \sum_{i=1}^{n} p_i q_i$ is the Lorentzian inner product. This formulation is unbounded and algebraically convenient, allowing closed-form computation of geodesic distances and stable gradient optimization. Because of its numerical robustness and simple analytical expressions for exponential and logarithmic maps, the Lorentz model is widely adopted in hyperbolic representation learning, particularly in entailment-based and hierarchical vision-language models.

\subsection{Value Function}
\label{sec:value}

Reinforcement learning (RL) is about learning to act in an environment so as to maximize future reward. 
The value function in RL estimates the expected cumulative future reward that an agent can obtain from a particular state or state-action pair \cite{sutton1998reinforcement}. It helps the agent decide which actions are more desirable in the long run, guiding it towards making decisions that maximize its total reward over time.
There are two main types: state-value function $V(s)$, which predicts the value of a given state, and state–action value function $V(s, a)$, which predicts the value of taking a specific action in a given state.

\paragraph{State Value Function.} The value function $V(s)$ represents the expected return (cumulative reward) starting from a specific state $s$ and following a particular policy, which defines the agent's strategy for choosing actions. It indicates how good it is for the agent to be in a certain state. For example, if an agent is located at a particular position in a maze, the value function $V(s)$ represents the expected total reward it will obtain from that point until it reaches the target, assuming it continues to follow its current policy.

\paragraph{State-Action Value Function.} The value function $V(s, a)$ or $Q(s, a)$, also known as the action value function, represents the expected return (cumulative reward) starting from a specific state $s$, taking an action $a$, and subsequently following a particular policy. It indicates how good it is for the agent to take a specific action in a given state. For example, if an agent is at a particular position in a maze, the function $V(s, a)$ represents the expected total reward it will obtain by choosing a particular action at that point and then following its current policy until it reaches the target.

\paragraph{Formal Definition.}
Given a state $s_t$, an action $a_t$, a policy $\pi(a \mid s)$ that defines how the agent acts, a reward function $r(s, a)$, and a discount factor $\gamma \in [0,1)$, the \textit{state-value function} under policy $\pi$ is formally defined as
\begin{equation}
V^{\pi}(s_t) = \mathbb{E}_{\pi} \left[ \sum_{k=0}^{\infty} \gamma^{k} \, r(s_{t+k}, a_{t+k}) \,\middle|\, s_t \right].
\end{equation}
It measures the expected cumulative reward that an agent will receive when starting from state $s_t$ and following policy $\pi$ thereafter.

\paragraph{Optimal Value Function.}
If the agent follows the best possible policy that maximizes the expected reward, we obtain the \textit{optimal value function}:
\begin{equation}
V^{*}(s_t) = \max_{\pi} \, \mathbb{E}_{\pi} \left[ \sum_{k=0}^{\infty} \gamma^{k} \, r(s_{t+k}, a_{t+k}) \,\middle|\, s_t \right].
\end{equation}
Intuitively, $V^{*}(s_t)$ quantifies the maximum expected cumulative reward that an agent can achieve when starting from state $s_t$ and following an optimal policy thereafter.

\paragraph{Bellman Optimality Equation.}
The value function encodes the long-term consequences of actions and forms the foundation of reasoning in reinforcement learning \cite{bellman1966dynamic}. Once $V^{*}(s)$ is known, the optimal policy can be derived if the one-step transition dynamics $P(s' \mid s, a)$ are available. Specifically, the optimal policy $\pi^{*}$ satisfies the Bellman optimality equation:
\begin{equation}
\pi^{*}(s) = \arg\max_{a} \left[ r(s, a) + \gamma \sum_{s'} P(s' \mid s, a) \, V^{*}(s') \right].
\end{equation}
This recursive relationship expresses how the value of a state depends on the values of its successor states, thereby capturing the essence of sequential decision-making.

\paragraph{Interpretation.}
The optimal value function $V^{*}(s)$ can be viewed as a potential field or energy map over the state space. States with high value correspond to desirable or low-energy configurations that are closer to reward, whereas states with low value represent undesirable or high-energy configurations that are further away. From this perspective, acting optimally can be interpreted as following the gradient of the value landscape toward regions of higher value (or lower energy). This analogy bridges reinforcement learning with energy-based modeling, suggesting that value functions implicitly define an energy surface that guides the agent toward optimal behavior.

\paragraph{Path Value Function.}

In generic RL settings above, the value depends on future rewards under the optimal policy. But if we redefine reward as the negative energy cost $-c$ of moving between states:

\begin{equation}
r(s, a, s') = -c(s, s'),
\end{equation}

and the cumulative reward becomes the total negative cost, then the optimal path value function $V^{*}(s, s')$ corresponds to the negative of the minimum accumulated cost from $s$ to $s'$ \cite{ciesielski2018path}.

In this case, the optimal path value function $V^{*}(s, s')$ obeys triangle inequality \cite{pitis2020inductive,csaji2008value,mitsuhashi2023triangle}:

\begin{align}
c(s_1, s_3) &\leq c(s_1, s_2) + c(s_2, s_3) \nonumber \\
\Rightarrow\quad V^{*}(s_1, s_3) &\geq V^{*}(s_1, s_2) + V^{*}(s_2, s_3).
\end{align}

\section{Motivation}

World state transitions (from video observations) naturally form a hierarchical structure that is suitable represented in hyperbolic space. Let $s_t$ denote the state at time $t$ and $\mathcal{A}$ be a discrete action set with cardinality $|\mathcal{A}| = B$. The world evolves according to the transition $s_{t+1} = f(s_t, a_t)$, where $a_t \in \mathcal{A}$. When predicting $d$ steps into the future, each action choice produces a distinct future trajectory, resulting in $N_d = B^d$ possible future states. These futures form a exponentially branching tree, where the depth corresponds to the prediction horizon and the branching factor is determined by the action space. As a result, future world states are naturally organized hierarchically: states at smaller depths represent coarse, high-level abstractions, while states at larger depths correspond to finer, more detailed futures. Similar motivations are also supported by~\cite{skenderi2025graph}.

\section{Baseline Details}
\label{sec:baseline}

As mentioned in Section~\ref{sec:baseline_protocol} of main content, in both Procedural Planning (PP) and Visual Planning with Videos setup, we evaluate against three categories of baselines. \textit{LLM-based} approaches rely on large language or vision-language models for reasoning, instruction following, and multi-step planning. \textit{Generative (world) models} perform planning by generating pixels or latent video tokens and using visual rollouts to guide decision-making. \textit{Predictive (world) models} focus purely on action prediction, estimating future action sequences directly without generating visual frames.

\textit{Random Selection.} Following prior work~\cite{chang2020procedure}, actions are sampled uniformly at random from the available action set to form a plan, without considering the task context.

\textit{Retrieval-Based.} Following prior work~\cite{zhou2025masked}, given the start and goal observations, this method retrieves the most similar trajectory from the training set by minimizing visual feature distance. The corresponding action sequence from the retrieved example is then used as the predicted plan.

\paragraph{LLM-based}

\textit{LFP (Language-First Planning)}~\cite{liu2023language}. This method first converts both the start and goal observations into text and then prompts a large language model to infer the missing steps. The LLM predicts a sequence of intermediate actions based solely on language reasoning rather than visual planning.

\textit{VidAssist}~\cite{islam2024propose}. This method uses a vision-language model to extract temporal and spatial cues from the video, then queries a large language model to interpret these cues and generate an action sequence. The LLM refines and structures the predicted steps into a coherent plan, combining visual grounding with language-based reasoning.

\textit{SCHEMA}~\cite{niu2024schema}. This method performs procedure planning by modeling how states evolve over time. It aligns visual observations with textual state descriptions through cross-modal contrastive learning and uses a transformer backbone to represent state transitions. A large language model is then used to reason over these inferred intermediate states and generate the next actions, enabling structured step-by-step planning in instructional video settings.

\textit{Other VLMs.} We also evaluate several large vision-language models, including InternVL3.5-241B~\cite{wang2025internvl3_5}, Qwen3-VL-Max~\cite{yang2025qwen3}, Gemini 2.5 Pro~\cite{comanici2025gemini}, and GPT-5~\cite{openai2025gpt5systemcard}, using them in a zero-shot setting to perform visual reasoning and planning directly from video observations without task-specific training.

\paragraph{Generative (World) Models}

\textit{DDN}~\cite{chang2020procedure}. This approach uses an autoregressive structure with two coordinated branches: one learns a compact representation of action steps, while the other predicts transitions in the latent feature space. By forecasting the next visual state rather than directly selecting actions, DDN models procedural progression through iterative frame prediction.

\textit{Int-MGAIL and Ext-MGAIL}~\cite{bi2021procedure}. These generative models perform procedure planning by jointly learning a latent world model and an action policy through adversarial training, enabling multi-step action synthesis conditioned on visual goal states.

\textit{P$^3$IV}~\cite{zhao2022p3iv}. This transformer-based model uses a learnable memory module together with an adversarial generation setup, and, similar to our method, outputs all action steps in a single forward pass rather than generating them sequentially.

\textit{PDPP}~\cite{wang2023pdpp}. This two-branch diffusion-based framework models temporal dependencies and action transitions, generating the full action sequence in parallel and progressively refining it over multiple denoising stages to improve coherence and logical structure.

\textit{KEPP}~\cite{nagasinghe2024not}. This method incorporates structured procedural knowledge through a probabilistic knowledge graph learned from training plans, which serves as external guidance for step ordering. KEPP predicts the full action sequence in a single pass with limited supervision, producing strong performance in instructional video planning.

\textit{ActionDiffusion}~\cite{shi2025actiondiffusion}. This diffusion-based approach generates the full action sequence by iteratively denoising a latent representation, allowing the model to refine predictions over multiple steps and capture long-term dependencies in instructional procedures.

\textit{MTID}~\cite{zhou2025masked}. This model treats procedure planning as a multimodal trajectory generation problem, using a diffusion-based latent policy to synthesize complete action sequences conditioned on video observations while modeling long-term dependencies through iterative denoising.

\textit{VideoWorld}~\cite{ren2025videoworld}. This autoregressive framework generates future video frames step by step to model procedural progression, using predicted visual states to implicitly guide the unfolding action sequence.

\paragraph{Predictive (World) Models}

\textit{WLTDO}~\cite{ehsani2018let}. This recurrent neural network model generates action sequences directly from paired observations, using temporal reasoning over the encoded features to predict ordered procedural steps.

\textit{UAAA}~\cite{abu2019uncertainty}. This two-stage method predicts action steps autoregressively by combining an RNN with a hidden Markov model to model temporal uncertainty and step transitions in procedural tasks.

\textit{UPN}~\cite{srinivas2018universal}. This method learns a differentiable latent space suitable for planning by predicting trajectories in feature space, and a softmax output layer is used to convert the continuous plan representation into discrete action steps.

\textit{PlaTe}~\cite{sun2022plate}. This model builds on DDN by introducing transformer modules into its dual-branch architecture for action and state prediction, but follows a distinct evaluation protocol compared to other procedure planning methods.

\textit{E3P}~\cite{wang2023event}. This method adopts an event-centric formulation, inferring latent events from visual observations and using them to guide intermediate action prediction. Through event-aware prompting and action relation modeling, E3P improves the logical structure of predicted steps and achieves strong performance on procedural planning benchmarks.

\textit{V-JEPA 2}~\cite{assran2025v}. A large-scale predictive world model pretrained on masked latent feature prediction over one million hours of unlabeled video. Action-conditioned post-training enables autoregressive rollouts for planning without pixel generation.

\section{Energy Landscape}
\label{sec:energy_landscape}

To better illustrate the difference between Euclidean predictive world models and our
hyperbolic formulation, we visualize the \emph{energy landscape} around a given latent
state.

\paragraph{$\Delta x$ and $\Delta y$.}

In the original V-JEPA 2-AC setup~\cite{assran2025v}, $\Delta x$ and $\Delta y$ represent physical end-effector offsets in Cartesian coordinates. The visualization shows how the model's energy changes as the end-effector's target position varies along the $\Delta x$ and $\Delta y$ axes while keeping the vertical displacement fixed ($\Delta z = 0$). 

Formally, the plotted quantity is:

\begin{align}
&s_{t+1}^{\mathrm{hyp}} = s_t + (\Delta x, \Delta y, 0),\\
&\text{Energy}(\Delta x, \Delta y) = c(s_t, s_{t+1}^{\mathrm{hyp}}).
\end{align}

where $c$ denotes the energy cost defined in Eq.~\ref{eq:energy_cost_small}.

In visual planning, $\Delta x$ and $\Delta y$ are no longer physical displacements. Instead, they represent latent displacements that probe the local geometry of the world model around a visual state.

In the Euclidean space, the encoder maps an observation $x_t$ into a latent vector
$s_t^{x} \in \mathbb{R}^n$. To visualize how the model evaluates hypothetical future
states, V-JEPA 2~\cite{assran2025v} perturbs the latent representation along two Euclidean axes. 
We choose two orthonormal directions in latent space, $u_1, u_2 \in \mathbb{R}^n$. A natural, semantically aligned choice is:

\begin{equation}
u_1 = \frac{E_\theta(x_{t+T}) - E_\theta(x_t)}
           {\left\lVert E_\theta(x_{t+T}) - E_\theta(x_t) \right\rVert},
\end{equation}

which represents the direction from the current state toward the goal (i.e., progress along the procedure). 

The second direction, $u_2$, spans variations orthogonal to this progress direction (i.e., sampled from another trajectory at the same step and then orthonormalized against $u_1$).

Then, a hypothetical next latent state is defined as

\begin{equation}
s^{\text{hyp}}_{t+1}
= s^x_t + \Delta x\, u_1 + \Delta y\, u_2,
\end{equation}

and the corresponding energy landscape is

\begin{equation}
\text{Energy}(\Delta x, \Delta y)
= \left\lVert s^x_{t+1} - s^{\text{hyp}}_{t+1} \right\rVert.
\end{equation}

In GeoWorld, the encoder maps each observation $x_t$ to a latent representation
$s_{t,\mathbb{H}}^{x}$ on the hyperbolic manifold. To probe the local geometry around
this latent state, we sweep two orthonormal directions in the tangent space
$\mathbf{T}_0\mathbb{H}^n$, denoted as $(\Delta x, \Delta y)$.
Each coordinate pair $(\Delta x, \Delta y)$ corresponds to a small displacement applied
\emph{at the tangent space} before projection onto the manifold via the exponential map:
\begin{equation}
s^{\text{hyp}}_{t+1,\mathbb{H}}
=
\exp_0\!\big(
s_t^{x}
+ \Delta x \, u_1
+ \Delta y \, u_2
\big),
\end{equation}
where $u_1$ and $u_2$ form an orthonormal basis in $\mathbf{T}_0\mathbb{H}^n$.
Thus, $(\Delta x,\Delta y)$ describes \emph{local perturbations of the latent state},
not pixel space offsets.

And the energy landscape is

\begin{equation}
\text{Energy}_{\mathbb{H}}(\Delta x, \Delta y)
= d_{\mathbb{H}}\!\left(s^{x}_{t+1,\mathbb{H}},\; s^{\text{hyp}}_{t+1,\mathbb{H}}\right).
\end{equation}

\paragraph{Visualization.}
In Figure~\ref{fig:energy_landscape}, we select a reference latent state $s_t$ from the initial step of the \textit{Replace Memory Chip} task in the COIN dataset~\cite{tang2019coin}, and visualize the local energy geometry by sweeping two orthonormal tangent-space directions $(\Delta x, \Delta y)$ around this state.
Figure~\ref{fig:energy_landscape} compares the Euclidean (left) and
hyperbolic (right) landscapes. The Euclidean surface shows a smooth, nearly symmetric
paraboloid with weak directional structure, indicating that V-JEPA~2 treats perturbations
homogeneously. In contrast, the hyperbolic surface in GeoWorld forms a sharper,
curvature-aware basin with more pronounced directional variation. This reflects the
ability of H-JEPA to encode hierarchical structure: states positioned higher in the task
hierarchy lie at hyperbolically greater distances, creating more informative energy
gradients during planning.

Such curvature-aware energy landscapes promote more stable long-horizon planning:
CEM naturally follows the hyperbolic geodesics shaped by GeoWorld, resulting in more
accurate multi-step trajectory optimization.

\begin{figure}[t]
    \centering
    \includegraphics[width=\linewidth]{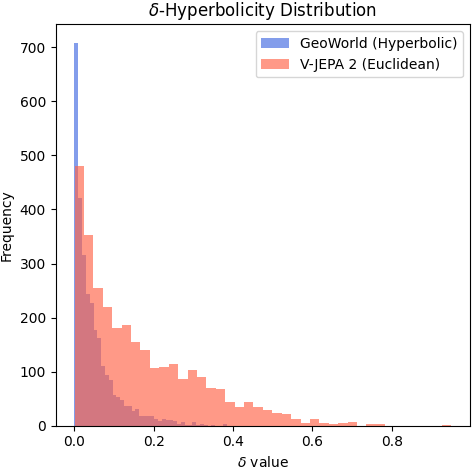}
    \caption{Gromov $\delta$-hyperbolicity on CrossTask~\cite{zhukov2019cross}.}
    \label{fig:hyperbolicity}
\end{figure}

\section{Ablation Study}
\label{sec:ablation}

\begin{figure*}[t]
    \centering
    
    \begin{subfigure}[t]{0.24\textwidth}
        \centering
        \includegraphics[width=\linewidth]{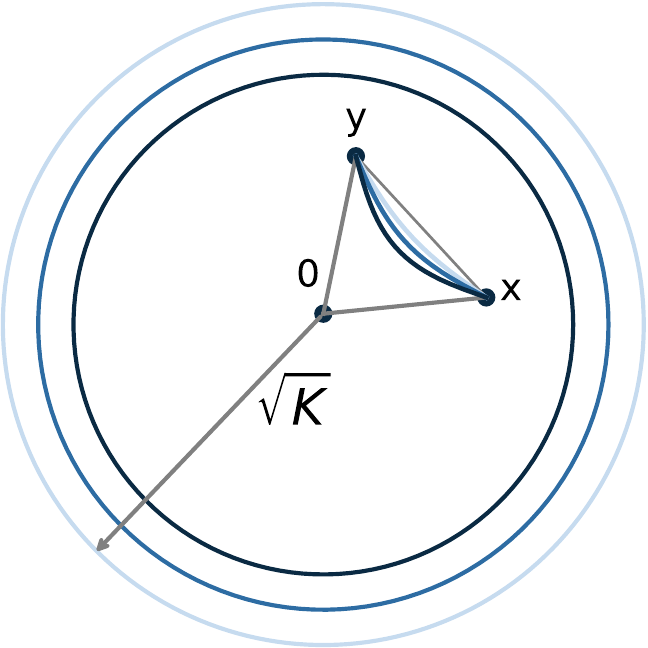}
        \caption*{(a) Curvature and geodesics.}
        \label{fig:disk}
    \end{subfigure}
    \hfill
    \begin{subfigure}[t]{0.24\textwidth}
        \centering
        \includegraphics[width=\linewidth]{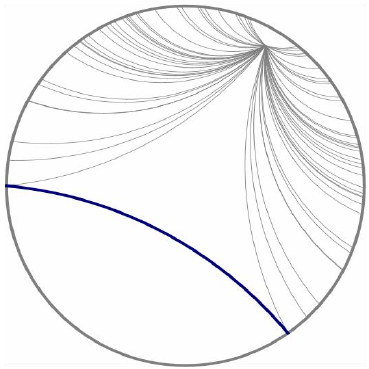}
        \caption*{(b) Geodesic patterns.}
        \label{fig:geodesic_lines}
    \end{subfigure}
    \hfill
    \begin{subfigure}[t]{0.24\textwidth}
        \centering
        \includegraphics[width=\linewidth]{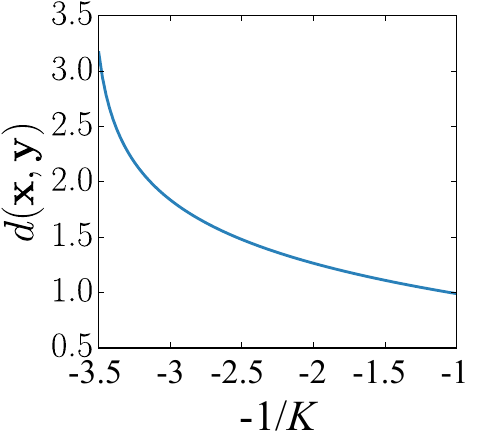}
        \caption*{(c) Distance vs. curvature.}
        \label{fig:curvature_geodesics}
    \end{subfigure}
    \hfill
    \begin{subfigure}[t]{0.24\textwidth}
        \centering
        \raisebox{2mm}{\includegraphics[width=\linewidth]{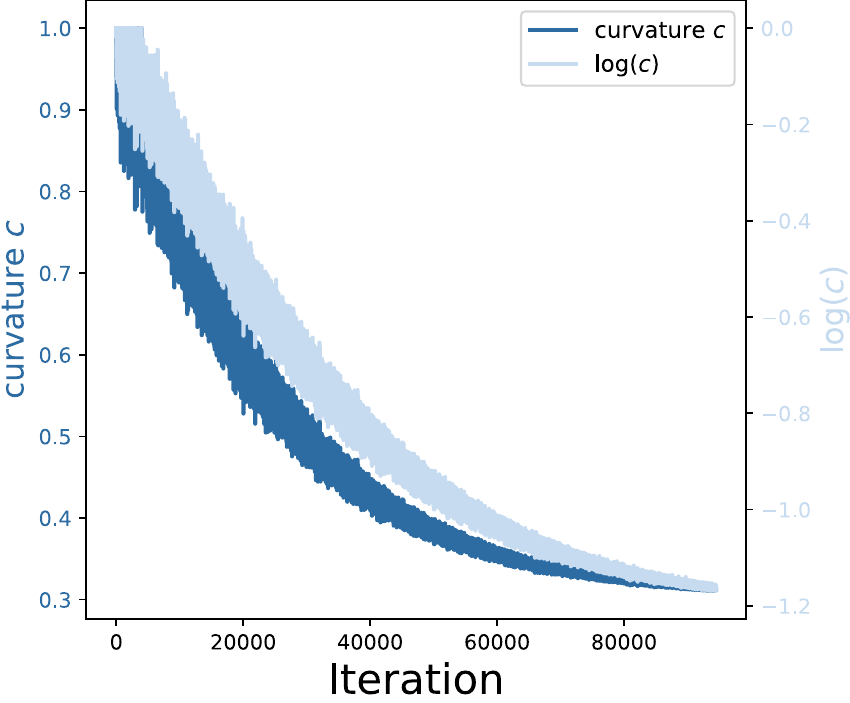}}
        \caption*{(d) Curvature trend during training.}
    \end{subfigure}
    \caption{\textbf{Geometric effects and curvature dynamics}: (a) Poincaré disk geodesics connecting $x$ and $y$ under different curvatures $K$. As the curvature $K$ becomes less negative (i.e., closer to $0$), the hyperbolic distance between $x$ and $y$ increases, and the geodesic paths bend less and shift closer toward the origin. (b) Geodesic patterns induced by different boundary anchor points. Varying the anchor location produces a characteristic geodesic fan in the Poincaré disk. (c) As the curvature becomes less negative, the space flattens and the distance between $x$ and $y$ decreases. (d) Learnable curvature $c$ during supervised training, showing a gradual decrease from its initialization and convergence to a stable value 0.3.}
    \label{fig:curvature}
\end{figure*}

\paragraph{Curvature}

As discussed in Section~\ref{sec:implementation}, the curvature \(K=-c\) is learned in the logarithmic space by optimizing \(\log(c)\), which is initialized at \(c=1\) and treated as a learnable scalar. This formulation ensures that \(c\) remains positive and stabilizes the gradients of both the hyperbolic distance and the exponential map~\cite{chami2019hyperbolic,desai2023hyperbolic}. The learned curvature is further clamped to the range \([0.1,\,10.0]\) to prevent training instability.

We analyze how the learnable curvature evolves during training and how it influences geometric planning quality. As shown in Fig.~\ref{fig:curvature}\,(d), the curvature parameter $c$ in GeoWorld typically starts near $1$ and gradually decreases to a stable value around $0.3$, indicating that the model learns a flatter yet still hyperbolic latent geometry. A smaller curvature reduces distortion in the exponential map and leads to more stable multi-step planning, especially for larger backbone encoders. The geometric effect of curvature is further visualized in Fig.~\ref{fig:curvature}\,(a)--(c): as $c$ decreases, geodesic paths bend less aggressively toward the origin (Fig.~\ref{fig:curvature}\,(a)), boundary-anchored geodesic patterns become flatter (Fig.~\ref{fig:curvature}\,(b)), and the hyperbolic distance between $x$ and $y$ contracts smoothly as curvature approaches zero (Fig.~\ref{fig:curvature}\,(c)). This suggests that moderate negative curvature is sufficient to capture hierarchical structure while preserving stable value propagation across long planning horizons.

\paragraph{Gromov $\delta$-Hyperbolicity}

We visualize Gromov $\delta$-hyperbolicity by sampling latent quadruples in CrossTask~\cite{zhukov2019cross} and evaluating the four-point condition under each model's intrinsic metric (hyperbolic geodesic distance for GeoWorld and Euclidean distance for V-JEPA 2). As shown in Fig.~\ref{fig:hyperbolicity}, GeoWorld exhibits a substantially more concentrated distribution of near-zero $\delta$ values, indicating a stronger tree-like hierarchical geometry in its learned representation space.

\paragraph{Frozen Encoder vs. Fully Fine-Tuned}

\begin{table}[t]
\centering
\small
\caption{Ablation of frozen encoder vs. fully fine-tuned model for visual planning with videos on CrossTask~\cite{zhukov2019cross}.}
\label{tab:ablation:frozen}
\resizebox{\linewidth}{!}{
\begin{tabular}{lcccccc}
\toprule
\multirow{2}{*}{Method}
& \multicolumn{3}{c}{T=3}
& \multicolumn{3}{c}{T=4} \\
\cmidrule(lr){2-4}\cmidrule(lr){5-7}
& SR & mAcc & mIoU
& SR & mAcc & mIoU \\
\midrule
GeoWorld ViT-L
& 44.80 & 70.54 & 86.30 & 30.63 & 65.46 & 79.73 \\
\rowcolor{cyan!20}w/ FFT
& 45.20 & 71.17 & 87.16 & 31.34 & 67.16 & 80.38 \\
GeoWorld ViT-H
& 47.79 & 74.42 & 88.84 & 34.51 & 68.89 & 82.95 \\
\rowcolor{cyan!20}w/ FFT
& 48.46 & 74.94 & 89.10 & 34.95 & 69.42 & 83.47 \\
GeoWorld ViT-g
& 49.23 & 76.64 & 90.61 & 35.49 & 71.00 & 84.50 \\
\rowcolor{cyan!20}w/ FFT
& 49.57 & 76.86 & 91.04 & 35.91 & 71.76 & 85.13 \\
GeoWorld ViT-g$_{384}$
& 51.71 & 77.30 & 92.95 & 37.04 & 71.35 & 87.04 \\
\rowcolor{cyan!20}w/ FFT
& 52.04 & 77.98 & 93.61 & 37.85 & 72.24 & 87.80 \\
\bottomrule
\end{tabular}
}
\end{table}

As shown in Table~\ref{tab:ablation:frozen}, we evaluate the impact of Fully Fine-Tuning (FFT) the encoder during the supervised finetuning stage, compared to the original configuration where the encoder remains frozen and only a lightweight exponential projection layer is trainable. Fully fine-tuning yields consistent yet modest improvements across all metrics and model scales, with gains of approximately \(0.3{-}0.8\%\) in SR and \(0.5{-}1.2\%\) in mAcc and mIoU for both \(T{=}3\) and \(T{=}4\) planning horizons. While these improvements indicate that the encoder can still adapt beneficially to downstream visual planning objectives, the gains come at the cost of significantly increased trainable parameters and slower optimization. Moreover, the relative performance margin narrows as model size increases, suggesting diminishing returns for larger backbones. These results imply that the frozen-encoder design already captures task-relevant structure effectively, and full encoder finetuning provides only incremental benefit relative to the additional computation and memory overhead introduced.

\begin{table}[t]
\centering
\small
\caption{Ablation of Supervised Fine-Tuning (SFT) vs. Geometric Reinforcement Learning (GRL) for visual planning with videos on CrossTask~\cite{zhukov2019cross}.}
\label{tab:ablation:GRL}
\resizebox{\linewidth}{!}{
\begin{tabular}{lcccccc}
\toprule
\multirow{2}{*}{Method}
& \multicolumn{3}{c}{T=3}
& \multicolumn{3}{c}{T=4} \\
\cmidrule(lr){2-4}\cmidrule(lr){5-7}
& SR & mAcc & mIoU
& SR & mAcc & mIoU \\
\midrule
V-JEPA 2 ViT-g$_{384}$~\cite{assran2025v}
& 50.16 & 74.86 & 91.73 & 35.01 & 70.24 & 85.05\\
\midrule
\multicolumn{6}{l}{\textit{GeoWorld ViT-g$_{384}$}} \\
SFT Only
& 50.42 & 75.13 & 91.94 & 35.92 & 70.79 & 85.88 \\
GRL Only
& 51.04 & 76.48 & 92.42 & 36.33 & 71.04 & 86.31 \\
\rowcolor{cyan!20}SFT + GRL
& 51.71 & 77.30 & 92.95 & 37.04 & 71.35 & 87.04 \\
\bottomrule
\end{tabular}
}
\end{table}

\begin{table}[t]
\centering
\small
\caption{Ablation of weighting hyperparameter $\lambda$ in Supervised Fine-Tuning (SFT) Only for visual planning with videos on CrossTask~\cite{zhukov2019cross}.}
\label{tab:ablation:SFT-hyperpara}
\resizebox{\linewidth}{!}{
\begin{tabular}{lcccccc}
\toprule
\multirow{2}{*}{Method}
& \multicolumn{3}{c}{T=3}
& \multicolumn{3}{c}{T=4} \\
\cmidrule(lr){2-4}\cmidrule(lr){5-7}
& SR & mAcc & mIoU
& SR & mAcc & mIoU \\
\midrule
V-JEPA 2 ViT-g$_{384}$~\cite{assran2025v}
& 50.16 & 74.86 & 91.73 & 35.01 & 70.24 & 85.05\\
\midrule
\multicolumn{6}{l}{\textit{GeoWorld ViT-g$_{384}$}} \\
$\lambda = 1,~ 1-\lambda = 0$
& 50.16 & 74.88 & 91.79 & 34.65 & 69.48 & 84.10 \\
$\lambda = 0.9,~ 1-\lambda = 0.1$
& 50.19 & 74.91 & 91.84 & 34.95 & 70.05 & 84.85 \\
$\lambda = 0.8,~ 1-\lambda = 0.2$
& 50.25 & 74.96 & 91.88 & 35.31 & 70.40 & 85.27 \\
$\lambda = 0.7,~ 1-\lambda = 0.3$
& 50.33 & 75.02 & 91.89 & 35.57 & 70.57 & 85.46 \\
$\lambda = 0.6,~ 1-\lambda = 0.4$
& 50.37 & 75.06 & 92.92 & 35.82 & 70.66 & 85.72 \\
\rowcolor{cyan!20}$\lambda = 0.5,~ 1-\lambda = 0.5$
& \textbf{50.42} & \textbf{75.13} & \textbf{91.94} & 35.92 & 70.79 & \textbf{85.88} \\
$\lambda = 0.3,~ 1-\lambda = 0.7$
& 50.39 & 75.07 & 91.82 & \textbf{35.97} & \textbf{70.86} & 85.74 \\
\bottomrule
\end{tabular}
}
\end{table}

\paragraph{Effectiveness of GRL}

As shown in Table~\ref{tab:ablation:GRL}, incorporating Geometric Reinforcement Learning (GRL) leads to clear and consistent improvements over the supervised fine-tuning (SFT) baseline. While SFT alone yields marginal gains over the pretrained V-JEPA 2 model, applying GRL independently further boosts SR, mAcc, and mIoU across both planning horizons, suggesting that GRL better aligns the learned energy landscape with multi-step planning objectives. The combination of SFT and GRL achieves the strongest performance, indicating that SFT provides a strong initialization while GRL refines the latent dynamics toward energy-minimizing trajectories required for long-horizon reasoning. These findings highlight the complementary nature of supervised learning and reinforcement-based value shaping in predictive world models.

\paragraph{SFT Hyperparameters}

As shown in Table~\ref{tab:ablation:SFT-hyperpara}, incorporating the rollout loss into SFT consistently improves visual planning performance over the pure one-step objective ($\lambda=1$). Once $1-\lambda > 0$, all metrics exhibit steady gains, indicating that multi-step rollout supervision provides additional temporal consistency beyond standard single-step training. 
As the rollout weight increases (i.e., smaller $\lambda$), improvements become more pronounced, particularly for the longer planning horizon ($T=4$). For example, SR and mIoU steadily increase as $\lambda$ decreases from $1$ to $0.5$, suggesting that stronger rollout supervision effectively mitigates error accumulation over longer sequences. This trend aligns with the intuition that longer-horizon prediction requires explicit multi-step consistency constraints rather than relying solely on local one-step accuracy.
A balanced weighting around $\lambda=0.5$ achieves the strongest overall performance across metrics, demonstrating that equal emphasis on one-step prediction and rollout consistency yields the best trade-off. Further increasing the rollout weight (e.g., $\lambda=0.3$) leads to negligible changes for the shorter horizon ($T=3$), while yielding slight yet consistent gains for the longer horizon ($T=4$). This behavior indicates that stronger rollout supervision primarily benefits long-horizon planning, especially under the hyperbolic structure where multi-step geodesic consistency becomes more critical. In contrast, short-horizon planning does not induce strong hierarchical structure, limiting the advantage of hyperbolic geometry and GRL. The primary benefit of GeoWorld emerges as the planning horizon increases, where exponential branching and long-term abstraction become critical, as shown in Table~\ref{tab:sr}.

\paragraph{GRL Hyperparameters}

As shown in Table~\ref{tab:ablation_GRL-hyperpara}, both the discount factor $\gamma$ and the regularization weight $\beta$ play important roles in shaping the learning dynamics in GRL. Increasing $\gamma$ strengthens long-horizon supervision by assigning greater weight to later predicted steps, which benefits multi-step rollout consistency and improves SR, mAcc, and mIoU as the planning horizon increases from $T{=}3$ to~$T{=}4$. Meanwhile, introducing the triangle inequality regularization term through $\beta > 0$ consistently boosts performance compared to the $\beta = 0$ setting, demonstrating that enforcing hyperbolic geodesic constraints helps stabilize the predictor and prevents degenerate shortcuts in latent space. Moderate regularization ($\beta = 0.1$) paired with a large discount factor ($\gamma = 0.99$) achieves the strongest results, indicating that encouraging long-horizon consistency while softly enforcing geodesic structure yields the most effective balance. These results validate the effectiveness of GRL as both a geometric constraint mechanism and a planning-aligned training signal.

\paragraph{Hyperbolic Geometry vs. GRL in Long-Horizon Planning}

\begin{table}[t]
\centering
\small
\caption{Ablation of discount factor $\gamma$ and the regularization weight $beta$ in Geometric Reinforcement Learning (GRL) for visual planning with videos on CrossTask~\cite{zhukov2019cross}.}
\label{tab:ablation_GRL-hyperpara}
\resizebox{\linewidth}{!}{
\begin{tabular}{lcccccc}
\toprule
\multirow{2}{*}{Method}
& \multicolumn{3}{c}{T=3}
& \multicolumn{3}{c}{T=4} \\
\cmidrule(lr){2-4}\cmidrule(lr){5-7}
& SR & mAcc & mIoU
& SR & mAcc & mIoU \\
\midrule
V-JEPA 2 ViT-g$_{384}$~\cite{assran2025v}
& 50.16 & 74.86 & 91.73 & 35.01 & 70.24 & 85.05\\
\midrule
\multicolumn{6}{l}{\textit{GeoWorld ViT-g$_{384}$}} \\
SFT Only
& 50.42 & 75.13 & 91.94 & 35.92 & 70.79 & 85.88 \\
$\beta = 0,~ \gamma = 0.99$
& 50.48 & 75.27 & 91.99 & 36.07 & 70.94 & 86.07 \\
$\beta = 0.05,~ \gamma = 0.99$
& 51.04 & 76.21 & 92.39 & 36.58 & 71.13 & 86.45 \\
$\beta = 0.2,~ \gamma = 0.99$
& 51.69 & 77.25 & 92.83 & \textbf{37.15} & 71.33 & 86.96 \\
$\beta = 0.1,~ \gamma = 0.90$
& 51.02 & 76.39 & 92.04 & 36.42 & 70.88 & 86.33 \\
$\beta = 0.1,~ \gamma = 0.95$
& 51.44 & 76.94 & 92.75 & 36.85 & 71.05 & 86.67 \\
\rowcolor{cyan!20}$\mathbf{\beta = 0.1,~ \gamma = 0.99}$
& \textbf{51.71} & \textbf{77.30} & \textbf{92.95} & 37.04 & \textbf{71.35} & \textbf{87.04} \\
\bottomrule
\end{tabular}
}
\end{table}

\textit{Section~\ref{sec:long}} in main paper reports results up to $T{=}6$, following the long-horizon setting in~\cite{zhou2025masked}. Table~\ref{tab:sr} further extends the evaluation to $T{=}8$ to stress-test planning stability under increasingly long rollouts. As the horizon grows, the vanilla V-JEPA~2 baseline exhibits rapid performance degradation, with SR dropping sharply from $50.16$ at $T{=}3$ to $4.95$ at $T{=}8$, highlighting severe error accumulation in long-horizon prediction.
Introducing hyperbolic geometry substantially mitigates this collapse. SFT in hyperbolic space already improves stability at longer horizons, maintaining significantly higher SR at $T{\geq}7$. Applying GRL in Euclidean space further strengthens multi-step consistency and consistently outperforms the baseline, demonstrating that rollout-based geometric regularization alone contributes meaningful gains even without hyperbolic modeling. 
When GRL is implemented in hyperbolic space, the advantage becomes more pronounced, particularly for $T{\geq}6$, suggesting that enforcing geodesic consistency in a curvature-aware latent space better preserves long-range structural dependencies. The full model (SFT + GRL) achieves the strongest results across all horizons, with the performance gap widening as $T$ increases. This trend indicates that SFT and GRL play complementary roles: SFT stabilizes short-term prediction, while GRL enhances long-horizon rollout consistency, together yielding a clear advantage in extended planning scenarios.

\begin{table}[t]
\centering
\small
\caption{SR of long horzion planning on CrossTask~\cite{zhukov2019cross} videos.}
\label{tab:sr}
\resizebox{\linewidth}{!}{
\begin{tabular}{lcccccc}
\toprule
Method & T=3 & T=4 & T=5 & T=6 & T=7 & T=8 \\
\midrule
V-JEPA 2 ViT-g$_{384}$
& 50.16 & 35.01 & 23.17 & 16.88 & 8.26 & 4.95 \\
SFT (Hyperbolic) & 50.42 & 35.92 & 23.64 & 16.97 & 14.88 & 11.51\\
GRL (Euclidean) & 50.26 & 35.47 & 23.85 & 17.03 & 15.12 & 12.74\\
GRL (Hyperbolic) & 51.04 & 36.33 & 24.05 & 17.82 & 15.54 & 13.10 \\
SFT + GRL
& 51.71 & 37.04 & 24.83 & 18.26 & 16.09 & 13.81\\
\bottomrule
\end{tabular}
}
\end{table}

\section{Error Accumulation in Long-Horizon Planning}

Autoregressive (AR) methods inevitably lead to error accumulation in long-horizon planning, which is why many existing works focus on mitigating this issue through rollout loss. However, our claim is not that hierarchy replaces this effect, but that geometry shapes how errors accumulate. In Euclidean latent spaces, small prediction errors cause unconstrained drift that compounds uniformly over time, whereas hyperbolic geometry imposes a hierarchical structure on the latent space that constrains long-horizon trajectories along geodesically meaningful directions. In this sense, error accumulation and geometric drift are closely related: hierarchical geometry mitigates how errors propagate, while rollout loss and GRL help eliminate them.

\section{Limitation and Future Work}

Our intuition for hierarchical structure arises from state transitions in multi-step planning over futures. Therefore, even when the action sequences annotated in CrossTask~\cite{zhukov2019cross} and COIN~\cite{tang2019coin} appear linear, predicting $d$-step futures from a state induces an exponentially branching set of possible trajectories ($B^d$), forming an implicit tree underlying a hierarchical structure. We must clarify that, as mentioned in \textit{Section~\ref{sec:hier}}, sub-task hierarchies involving multi-level planning are the intuition of the original JEPA~\cite{lecun2022path}. However, the hierarchical structure in GeoWorld arises from multi-step future expansion, rather than from explicit high-level planning and low-level execution.

Future work may involve sub-task hierarchies, such as high-level task labels, mid-level actions, and low-level end-effectors. Moreover, our framework is compatible with embodied planning. As this is a computer vision conference, we plan to extend our work to embodied settings in the future.

\end{document}